%% file: main.tex
\documentclass[letterpaper, 10 pt, journal, twoside]{IEEEtran}
\usepackage{times}
\usepackage{amsmath,amsfonts,amssymb,amsthm,bm}
\usepackage{graphicx}
\usepackage{graphicx}
\usepackage{subcaption}
\usepackage{cancel}
\usepackage{comment}
\usepackage[linesnumbered,ruled, noend]{algorithm2e}
\usepackage{multicol}
\usepackage[bookmarks=true]{hyperref}
\usepackage{multirow}
\usepackage[table]{xcolor}
\usepackage{acronym}
\usepackage{xspace}
\usepackage{mathtools}
\usepackage{siunitx}
\usepackage{comment}
\usepackage[normalem]{ulem}
\usepackage{tikz}
\usepackage{svg}
\usepackage{pbalance}
\usepackage{booktabs}

\IEEEoverridecommandlockouts                              

\input{commands.tex}

\title{
Curve-Induced Dynamical Systems on \\ Riemannian Manifolds and Lie Groups
}

\author{
Saray Bakker$^{*1}$, Martin Schonger$^{*2, 3}$, Tobias L{\"o}w$^{2, 3}$, Javier Alonso-Mora$^{1}$ and Sylvain Calinon$^{2, 3}$
\thanks{*These authors contributed equally and are mentioned in alphabetic order.}
\thanks{The authors are associated with $1.$ Delft University of Technology, The Netherlands, $2.$ Idiap Research Institute, Switzerland, and $3.$ \'{E}cole Polytechnique F\'{e}d\'{e}rale de Lausanne (EPFL), Switzerland.
}
\thanks{This project has received funding from the European Union through ERC, INTERACT, under Grant 101041863, and the Swiss National Science Foundation (SNSF) through the HORACE project (Grant 10002813, \href{https://horace-robotics.github.io}{https://horace-robotics.github.io}). Views and opinions expressed are those of the author(s) and do not necessarily reflect those of the granting authority. Neither the European Union nor the granting authority can be held responsible for them.}
\thanks{Project website and video: \href{https://cd-sm.github.io/}{https://cd-sm.github.io}.
}
\thanks{Corresponding authors: Saray Bakker (\textit{s.bakker-7@tudelft.nl}) and Martin Schonger (\textit{martin.schonger@epfl.ch}).}
}

\begin{document}

\maketitle


\setlength{\abovedisplayskip}{1.5ex plus3pt minus1pt}
\setlength{\belowdisplayskip}{\abovedisplayskip}
\captionsetup{font=small}

\input{sections/0_abstract}
\input{sections/1_introduction}
\input{sections/2_related_works}
\input{sections/3_preliminaries}
\input{sections/4_methods}
\input{sections/4b_relevant_manifolds}
\input{sections/5_results}
\input{sections/7_conclusion}
\appendices
\input{sections/8_appendix}

\bibliographystyle{IEEEtran}
\bibliography{references}

\end{document}

%% file: commands.tex
\newcommand{\norm}[1]{\left\lVert#1\right\rVert}

\newcommand{\N}{\mathbb{N}}

\newcommand{\R}{\mathbb{R}}

\definecolor{OliveGreen}{rgb}{0,0.6,0}

\acrodef{nlp}[NLP]{nonlinear program}
\acrodef{dof}[DoF]{degrees of freedom}
\acrodef{qp}[QP]{quadratic program}
\acrodef{if}[G3F]{Globally-Guided Geometric Fabrics}
\acrodef{gf}[GF]{Geometric Fabrics}
\acrodef{rf}[RF]{Rollout Fabrics}
\acrodef{prf}[P-RF]{Prioritized Rollout Fabrics}
\acrodef{tamp}[TAMP]{Task and Motion Planning}
\acrodef{mpc}[MPC]{Model Predictive Control}
\acrodef{spd}[SPD]{symmetric positive definite}
\acrodef{mp}[MP]{motion primitive}
\acrodef{mps}[MPs]{motion primitives}
\acrodef{ds}[DS]{dynamical system}
\acrodef{dsmp}[DS-MP]{DS-MP}
\acrodef{dsmpman}[CDSM]{Curve-induced Dynamical systems on Smooth Manifolds}

\newcommand{\ie}{i.e.\@\xspace}
\newcommand{\eg}{e.g.\@\xspace}

\theoremstyle{definition}
\newtheorem{definition}{Definition}[section]

\theoremstyle{remark}

\DeclareMathOperator*{\argmin}{argmin}

\newcommand{\transpose}{\mathsf{T}}

\newcommand{\PT}[2]{\mathcal{P}_{#1 \to #2}}  

\newcommand{\gfrak}{\mathfrak{g}}  
\newcommand{\ip}[2]{\langle #1, #2 \rangle}  

\newcommand{\Hspace}{\mathbb{H}}  
\newcommand{\UQ}{\mathbb{H}_1}  
\newcommand{\UQR}{\ensuremath{\mathbb{H}_1 \ltimes \mathbb{R}^3}}  
\newcommand{\se}{\mathfrak{se}(3)}  

\newcommand{\Sym}{\mathcal{S}^n}  
\newcommand{\SPDn}{\mathcal{S}_{++}^n}  
\newcommand{\SPD}[1]{\mathcal{S}_{++}^{#1}}  

\newcommand{\proj}{\pi}  

\DeclareSIUnit{\rad}{rad}

\newcommand{\Bezier}{B\'ezier\xspace}

\newcommand{\added}[1]{{\color{black}#1}}


%% file: sections/0_abstract.tex
\begin{abstract}
Deploying robots in household environments requires safe, adaptable, and interpretable behaviors that respect the geometric structure of tasks.
Often represented on Lie groups and Riemannian manifolds, this includes poses on $SE(3)$ or symmetric positive definite matrices encoding stiffness or damping.
In this context, dynamical system-based approaches offer a natural framework for generating such behavior, providing stability and convergence while remaining responsive to changes in the environment.
We introduce \ac{dsmpman},
a real-time framework for constructing \added{and coupling motion profiles} directly on Riemannian manifolds and Lie groups. The proposed approach constructs
a nominal curve on the manifold, and generates a dynamical system which combines a tangential component that drives motion along the curve and a normal component that attracts the state toward the curve.
We provide a stability analysis of the resulting dynamical system and validate the method quantitatively. 
\added{
We leverage the phase, indicating the progression along the curve, firstly, to define a phase modulation layer, that can modulate the timing while keeping a fixed spatial profile, and secondly, to couple dynamical systems and curves on different manifolds. We showcase the coupling's applicability in robot-assisted shoulder rehabilitation, and a real-world dressing task on both a robotic manipulator, where poses on $SE(3)$ and damping matrices on $\SPDn$ are adapted online, and on a mobile manipulator. 
In quantitative evaluations on $S^2$ and in robot-assistive tasks, \ac{dsmpman} demonstrates its effectiveness while achieving real-time curve-generation and fast query times.
}
\vspace{0mm}

\end{abstract}

%% file: sections/1_introduction.tex
\section{Introduction} \label{sec:introduction}
\begin{figure}
\centering
\vspace{1mm}
\includegraphics[width=\linewidth]{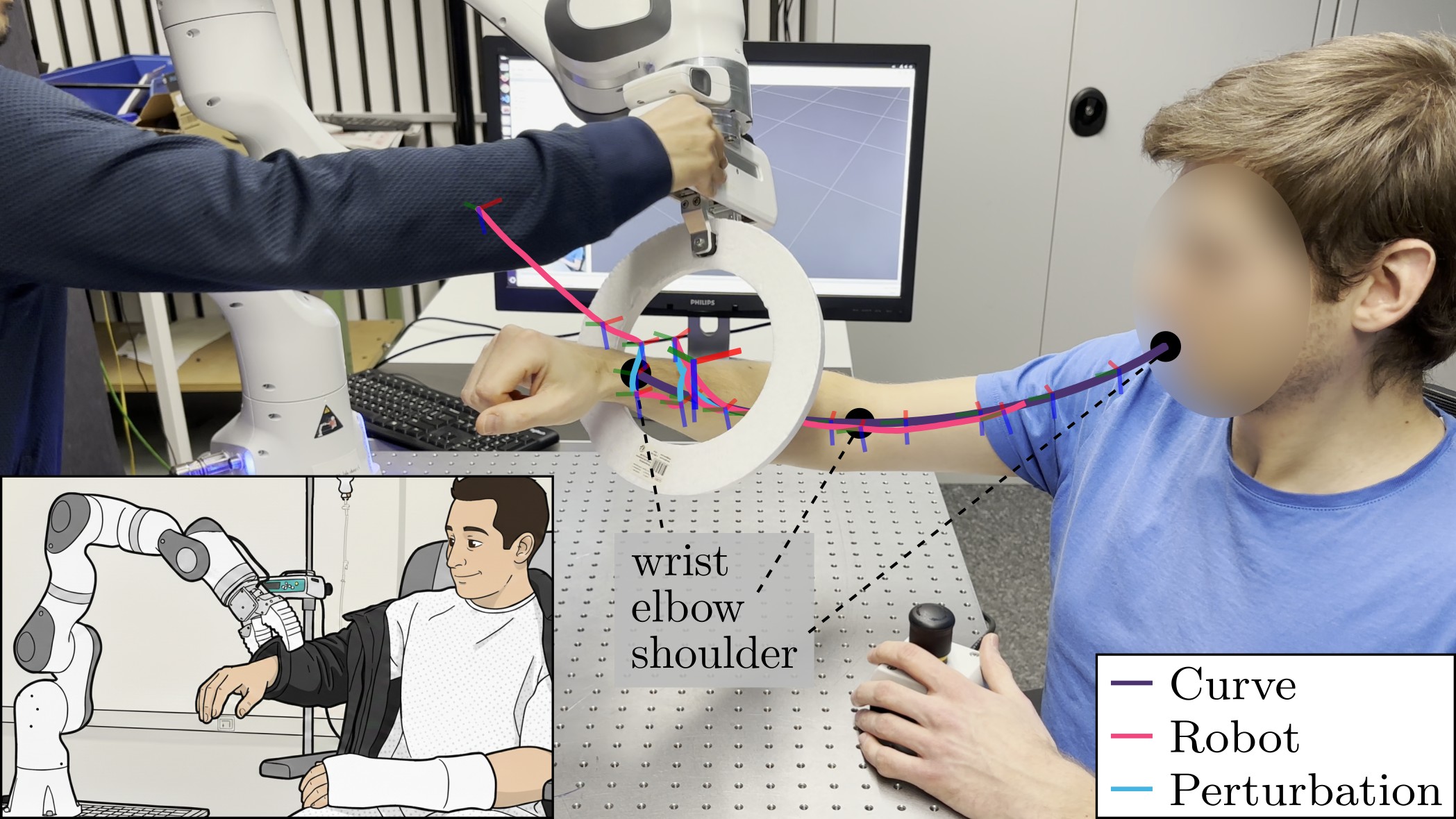}
\caption{%
We consider a dressing task in which the robot adapts both its pose \added{on $SE(3)$} and the \added{affine transformation for the desired velocity} of an impedance controller throughout the motion. External perturbations are applied by physically pushing the robot, and we observe that the proposed \ac{dsmpman} converges back to the desired nominal trajectory, given the tracked configuration of the human arm via a camera.
The illustration in the bottom-left corner was generated using Google Gemini.
}
\label{fig:dressing}
\end{figure}

Real-world deployment of robots in household settings requires systems that are inherently safe, adaptable, and interpretable. Achieving such safe and reliable behavior requires models that respect the underlying structure of the complex tasks these robots perform. In many cases, the processed data is inherently non-Euclidean, residing on curved manifolds rather than in \added{flat} vector spaces.
Considering the geometry of the task, such as poses on $SE(3)$, or stiffness and damping expressed as \acf{spd} matrices on $\SPDn$, is crucial for effective task execution as well as safety during interaction~\added{\cite{ManifoldBook, Calinon26AR}}.
For instance, in a robot-assisted dressing task, as shown in Fig.~\ref{fig:dressing}, the robot adapts its end-effector pose \added{on $SE(3)$} to the human arm's configuration and varies the affine transformation for the desired velocity in the impedance controller, throughout the interaction.

Dynamical system-based approaches offer an appealing framework for generating robot behavior in such settings, as they can provide formal guarantees on stability and convergence while remaining highly reactive to changes in the environment~\added{\cite{khansari2011lasa, billard2022learning}}. Several approaches have been developed, ranging from learning-based to deterministic methods.

Recently, there has been growing interest in learning stable \acp{ds} on Riemannian manifolds and Lie groups.
A prominent line of work leverages normalizing flows as diffeomorphic mappings, both for Lie groups~\cite{urain2022learning} and for more general Riemannian manifolds~\cite{zhangLearningRiemannianStable2023}. Closely related approaches employ Flow Matching on Riemannian manifolds~\cite{ding_fast_2024, chenFlowMatchingGeneral2024}
or enforce stability explicitly through constraints in the loss function~\cite{perez2024puma}.
Broadly speaking, data-driven methods still make specific assumptions to achieve stability that could limit performance on manifolds. They typically require several demonstrations, and training times hinder real-time adaptation of the learned DS to changes in the environment or task, such as adapting to different human arm configurations in the dressing task. 
In addition, convergence in a learned \ac{ds} is typically directed toward the goal, rather than the demonstrated trajectories, even though explicitly attracting the system to those trajectories could improve user-interpretability and task execution, especially when working with a small dataset.

This motivates the use of vector-field-based approaches defining a reference curve and constructing a vector field that draws motion toward this curve while also steering it along the curve's direction. These curves can be constructed as \ac{mps}, an effective and lightweight approach to encode a trajectory or data sequence, which can be defined on Riemannian manifolds and Lie groups~\cite{mancinelliSplinesManifoldsSurvey2024, wattersonTrajectoryOptimizationManifolds2020}. In Euclidean space, Li and Calinon~\cite{Li25arXiv2} show that an analytic distance map can be generated when using composite quadratic \Bezier curves, and converted into a \added{\ac{ds}} to react to disturbances. 
\added{Extensions to Riemannian manifolds have been proposed, but they inherit the way in which the reference is represented. Describing the desired path as an intersection of zero-level sets remains nontrivial for arbitrary demonstrated trajectories~\cite{yaoGuidingVectorFields2023}, whereas~\cite{lee2025behavior} proposes a vector field toward a discrete demonstration, and learned version that captures multiple learned behaviors in a continuous \ac{ds} representation.}

\added{We propose \acf{dsmpman}, a framework for generating and coordinating robot motion profiles on Riemannian manifolds and Lie groups. Building on the Euclidean-space DS in~\cite{Li25arXiv2}, we represent demonstrated motions by smooth phase-parametrized curves on Riemannian manifolds and
Lie groups. From this representation, we construct a continuous-time \ac{ds} for reactive motion generation on the manifold, combining an attracting component that guides the state toward the curve with a tangential component that propagates the motion along the direction of the curve.}
\added{Beyond providing a smooth reference, the parametrization turns the phase into a state-derived coordinate: through its projection onto the curve, every state in the analysis domain is assigned a phase. We exploit this coordinate in three ways. 
First, the smooth reference and phase can be used to show local convergence of the \ac{ds} toward the curve and the goal, as well as be used to show stability in case trajectories cross the ridges of the distance potential, where the closest point projection is non-unique.
Second, we introduce a phase modulation layer that allows the timing of a motion to be re-specified independently of its geometry, enabling, for example, adaptation to velocity limits and dynamic obstacles or agents while preserving the spatial curve and its stability properties.}

\added{Lastly, we propose a phase-based coupling of quantities defined on different manifolds. 
This is particularly relevant for human-robot interaction, where both the generated trajectory and the variation of stiffness and damping throughout the task are important. 
In contrast to approaches that jointly encode motion and impedance behavior in a single potential function~\cite{khansari2017learning}, \ac{dsmpman} provides a phase-based mechanism to couple \ac{ds}s and curves defined independently on different manifolds.
In particular, stiffness and damping matrices of an impedance controller can be defined as curves of full $\SPD{6}$-matrices~\cite{ajoudani2015reduced}, enabling extensions beyond approaches that leverage \emph{diagonal} stiffness and damping matrices~\cite{chenClosedLoopVariableStiffness2021, geigerDiffusionBasedImpedanceLearning2025}.}
\added{This phase-based coupling is demonstrated in an arm-dressing task by synchronizing an $SE(3)$ pose \ac{ds} with a curve of full $\SPD{6}$ matrices of the impedance controller, and in robot-assisted shoulder rehabilitation by coupling an $S^2$ shoulder-position \ac{ds} with velocity-gain matrices in $\SPD{2}$ for adaptive assistance throughout the exercise.}

\added{We summarize our contributions as follows:}
\begin{itemize}
    \item A curve-induced \added{\ac{ds}-formulation on} Riemannian manifolds\added{, with} an extension to Lie groups, \added{induced by a phase-parametrized smooth curve that is constructed from one or multiple} demonstrated trajectories or data sequences in real-time.
    \item \added{A stability analysis of the proposed \ac{ds}, establishing local convergence to the curve and goal, and analyzing stability for trajectories crossing the distance-potential ridges.}
    \item \added{A phase modulation layer that re-specifies the timing of the motion independently of its geometry, leveraging the same parametric curve structure, while leaving both the spatial curve and the stability analysis unchanged.}
    \item \added{The phase of the curve as a synchronization interface between manifolds, coupling \ac{ds}s and curves on different manifolds.
    We demonstrate this phase-dependent coupling for arm dressing by synchronizing a \ac{ds} of $SE(3)$ poses with varying $\SPD{6}$ encoding the affine transformation for the desired velocity in the impedance controller, and for shoulder rehabilitation by coupling $S^2$ exercises with assist-as-needed velocity gains on $\SPD{2}$.}
\end{itemize}
In quantitative experiments, our approach is compared against \added{learning-based and deterministic dynamical systems suitable for manifolds,} on the LASA dataset mapped to $S^2$. \added{The applicability of the coupling between \ac{ds}s and curves on different manifolds, is shown via a robot-assisted shoulder rehabilitation example, and} real-world deployment for a human-arm dressing task on a manipulator, as well as on a mobile manipulator.
\added{We provide fast curve optimization and batch evaluation of the \ac{ds} using JAX~\cite{jax2018github, jaxopt_implicit_diff}, enabling real-time queries and online curve updates, as demonstrated for dressing a moving human arm and in the quantitative comparison on $S^2$.
Most examples include manifolds on which geodesics, distances, and parallel transport are available analytically or can be evaluated efficiently, i.e. $S^2$, $SO(3)$, $SE(3)$, and $\SPDn$ up to $n=6$. We additionally provide an illustrative example for a Riemannian manifold where these expressions are not available analytically.}
Our code will be released open-source upon publication.

%% file: sections/2_related_works.tex
\section{Related Work}\label{sec:related_works}
In this section, we analyze several DS-based approaches for generating stable motions on smooth manifolds, ranging from deep-learning based approaches to guiding vector fields.

\subsection{Learning Stable Dynamical Systems on Manifolds}
Recently, several methodologies have been proposed for incorporating safety and stability guarantees leveraging learned \added{\ac{ds}s}~\cite{hu2024fusion}. 
By leveraging Neural ODEs, normalizing flows learn a diffeomorphic mapping between the observation space and latent space to generate a stable vector field, ensuring stability by connecting the task-space system to a simple, stable latent-space \added{\ac{ds}}. The method has been generalized to Lie groups~\cite{urain2022learning} and Riemannian manifolds~\cite{zhangLearningRiemannianStable2023}.
Flow Matching~\cite{lipman2022flowmatching} trains a Continuous Normalizing Flow resulting in a simplified procedure and better scalability compared to the likelihood maximization used in normalizing flows, and has been extended to several types of Riemannian manifolds~\cite{ding_fast_2024, chenFlowMatchingGeneral2024}.
Learning a bijective mapping to a simple hand-defined system in the latent space can be restrictive especially for non-Euclidean data. By enforcing stability of the DS via an additional term in the loss function, P\'erez-Dattari et al.~\cite{perez2024puma} avoid mapping to a simple system, and showcase its performance on $S^2$ and $SE(3)$-manifolds. 
Diffusion policies, on the other hand, have the potential to handle multimodal distributions, and are demonstrated on $SE(3)$ in~\cite{urain_se3_2022}. However, diffusion policies do not provide formal safety and stability guarantees. 

All learning-based methods mentioned above require a training procedure that typically ranges from several minutes to hours. By contrast, small-scale parametric models such as Gaussian Mixture Models (GMMs) can learn diffeomorphic mappings for motions on Riemannian manifolds~\cite{saverianoLearningStableRobotic2023}, with training procedures that can be executed in real time. However, ensuring that the learned diffeomorphism is sufficiently expressive, bijective, and smoothly invertible remains challenging, and often imposes restrictive design constraints, particularly when such mappings are applied on Riemannian manifolds.

\subsection{Dynamical Systems on Non-Euclidean Spaces}
Deterministic trajectory generation methods on smooth manifolds that do not rely on learning-based pipelines have been extensively studied. This includes work on second-order nonlinear oscillators defined directly on manifolds~\cite{fioriSyntheticNonlinearSecondorder2022}, as well as Dynamic Motion Primitives (DMPs) adapted to non-Euclidean spaces. In particular, DMP formulations have been proposed for orientations using unit quaternions~\cite{koutras2020correct, ude2014orientation}, for poses~\cite{saveriano2019merging}, and more generally for Riemannian manifolds, including the space of $\SPDn$-matrices~\cite{abu-dakkaUnifiedFormulationGeometryaware2024}. A probabilistic extension of DMPs for orientations represented by unit quaternions was introduced in~\cite{rozo2022orientationPDMP}.
These second-order nonlinear oscillators and DMPs primarily focus on reproducing a single nominal trajectory on the manifold. \added{Accurate trajectory generation and well-defined behavior far away from the demonstrated or prescribed trajectory are thereby not addressed simultaneously.}

To enable robust and safe motion generation from off-trajectory conditions, vector fields and \added{\ac{ds}s} are proposed, including guiding vector fields~\cite{goncalves2010vector, wu2018guidance}, velocity fields for contour following~\cite{li2002passive}, and stable \added{\ac{ds}s}~\cite{Li25arXiv2, figueroa2022locally}. These approaches typically define an action by combining an attraction term toward the curve with a continuation term following the movement profile of the demonstrated trajectory or movement toward the global goal.
However, such formulations are generally developed in Euclidean spaces. When naively extended to Riemannian manifolds or Lie groups via a single tangent-space projection, they can lead to significant geometric distortions~\cite{jaquier_unraveling_2024}.

In~\cite{yaoGuidingVectorFields2023}, a convergence and stability analysis of guiding vector fields on Riemannian manifolds is provided. Given an $n$-dimensional configuration space, their approach constructs a reference path as an intersection of $(n-1)$-dimensional zero-level sets, to which the vector field should converge.
Although they thoroughly analyze the difficulties of achieving global asymptotic stability for curves homeomorphic to the unit circle, their focus on examples with hand-defined and elementary curves limits the direct applicability of their results to the complex geometric structures encountered in robotic tasks.
For an arbitrary demonstrated trajectory, it remains nontrivial to define the desired path as an intersection of zero-level sets, and it becomes challenging to separate the attraction and progression component with respect to the original path, from the vector field directing toward the zero-level sets.

\added{Most closely related to our \ac{ds} is the guidance vector field of~\cite{lee2025behavior}, which combines a contracting velocity toward the closest point on a reference with a parallel-transported mimicking velocity and weights the two components. Instead, our distinction lies in representing the reference as a smooth, phase-parametrized curve fitted to one or more demonstrations, rather than directly using a finite sequence of demonstrated configurations. This continuous representation allows the reference, its phase, and its velocity to be evaluated independently of the sampling density of the demonstrations and enables us to construct a continuous-time \ac{ds} without learning a vector-field approximation. More importantly, the explicit phase provides a common coordinate that we exploit to analyze convergence and stability, including crossing ridges where the closest-point projection is non-unique, to coordinate \ac{ds}s and curves on different manifolds, and to independently modulate the timing of the motion.}

%% file: sections/3_preliminaries.tex
\section{Preliminaries} \label{sec: preliminaries}
\subsection{Riemannian Manifolds}
\added{Several quantities in robotics can be naturally represented on smooth manifolds, see~\cite{Calinon26AR} for a review and~\cite{ManifoldBook} for application examples. In control applications, stiffness and damping can be represented as symmetric positive definite matrices.}
Let $M$ be a smooth manifold of dimension $n$. Locally, the manifold resembles the Euclidean space $\R^n$, but globally it may have a complex topological structure.
At every point $p \in M$, we associate an $n$-dimensional vector space $T_p M$ called the \emph{tangent space}. 
Intuitively, this space consists of all possible tangent vectors to curves passing through $p$.
Formally, if $\gamma \colon (- \epsilon, \epsilon) \to M$ is a smooth curve with $\epsilon > 0$ and $\gamma(0) = p$, the velocity vector $\dot{\gamma}(0)$ is an element of $T_p M$ \cite{leeIntroductionSmoothManifolds2013}.
The collection of all tangent spaces forms the \emph{tangent bundle}, denoted $TM = \bigsqcup_{p \in M} T_p M$. A \emph{vector field} $X$ on $M$ is a smooth section of this bundle; it assigns a tangent vector $X_p \in T_p M$ to every point $p$ in a smooth manner.

A \emph{Riemannian manifold} is a pair $(M, g)$, where $g$ is a \emph{Riemannian metric}. When $g$ is clear from context, we simply refer to $M$ \added{as the Riemannian manifold}.
The metric $g$ is a smooth field of inner products, assigning to each $p \in M$ a symmetric, positive-definite bilinear form $\ip{\cdot}{\cdot}_p \colon T_p M \times T_p M \to \R$.
This allows us to define the length of tangent vectors as $\norm{v}_p = \sqrt{\ip{v}{v}_p}$.

\added{Tangent vectors at different points belong to different vector spaces and cannot generally be compared directly.}
The metric induces a unique torsion-free affine connection $\nabla$, called the \emph{Levi-Civita connection} \cite{leeIntroductionRiemannianManifolds2018}.
This connection allows us to differentiate vector fields and compare tangent vectors at different points.
Given a curve $\gamma \colon [t_0, t_1] \to M$, the connection defines a linear isometry known as \emph{parallel transport}, which preserves inner products along the curve. \added{Intuitively, it carries a tangent vector along the curve without changing its length.}
A \emph{geodesic} is a locally length-minimizing curve $\gamma$ with zero \added{covariant} acceleration. \added{It is the Riemannian analog of a straight line.}
For parallel transport along the \added{minimizing} geodesic connecting $p \in M$ to $q \in M$\added{, assumed unique}, we use the notation:
\begin{equation}\label{eq:parallel_transport}
    \PT{p}{q} \colon T_{p} M \to T_{q} M.
\end{equation}

\added{The exponential and logarithmic maps provide a local correspondence between a Riemannian manifold and its tangent spaces.} The \emph{exponential map},
\begin{equation}
    \exp_p \colon T_p M \to M,
\end{equation}
maps a tangent vector $v \in T_p M$ to the point on the manifold reached by the geodesic starting at $p$ with velocity $v$ after one unit of time. \added{Thus, $\exp_p$ can be viewed as the manifold counterpart of the Euclidean operation $p+v$.}
The \emph{logarithmic map},
\begin{equation}
    \log_p \colon M \to T_p M,
\end{equation}
is the \emph{local} inverse of the exponential map. It maps a point $q \in M$ to the tangent vector in $T_p M$ representing the initial velocity of the geodesic connecting $p$ to $q$. \added{It is the counterpart of the Euclidean difference $q-p$.}

The \emph{Riemannian distance}, or geodesic distance, \added{between $p$ and $q$} is defined intrinsically via the logarithm:
\begin{equation}\label{eq:RM_distance}
    d_M(p, q) = \norm{\log_p(q)}_p.
\end{equation}
\added{Concrete instances of these concepts for the $\SPD{n}$ manifold are introduced in Sec.~\ref{sec:spd}.}

\subsection{Lie Groups}
A \emph{Lie group} $G$ is a smooth manifold that is also a group in the algebraic sense, with smooth multiplication, $m \colon G \times G \to G, (g,h) \mapsto gh$, and inverse maps, $i \colon G \to G, g \mapsto g^{-1}$, respectively~\cite{sola_micro_2021}.
\added{A prominent example in robotics is the rigid-body transformation group $SE(3)$, which represents the pose of a rigid body.}

\added{The group structure also relates the tangent spaces at different elements of $G$, which is what allows all of them to be described by a single one.}
For any fixed element $g \in G$, the \emph{left translation} map $L_g \colon G \to G$ is defined by $L_g(h) = gh$\added{, where $h \in G$}.
Since $G$ is a smooth manifold, we can consider the differential of this map, often called the \emph{pushforward}, at any point $h \in G$.
We denote this pushforward by $d(L_g)_h \colon T_h G \to T_{gh} G$.
It is a linear map that \added{takes} tangent vectors from the tangent space at $h$ to the tangent space at $gh$.

A vector field $X$ is called \emph{left-invariant} if it is preserved by the pushforward of left translations. Formally, for all $g, h \in G$, the field must satisfy $d(L_g)_h (X_h) = X_{gh}$.
Here, $X_h$ denotes the value of the vector field $X$ evaluated at point $h$. The condition states that if we take the vector at $h$ and push it forward by left-multiplying by $g$, we obtain exactly the vector assigned by the field at the point $gh$.

The \emph{Lie algebra} $\gfrak$ of $G$ is the space of all left-invariant vector fields on $G$, equipped with the Lie bracket operation $[X, Y]$.
As a vector space, it is standard to identify $\gfrak$ with the tangent space at the identity, $T_e G$\added{, through evaluation at $e$. An element of $\gfrak$ can therefore be handled as a single tangent vector at the identity, which is how we use it throughout}.
The \emph{Lie group exponential},
\begin{equation}
    \exp \colon \gfrak \to G,
\end{equation}
maps lines \added{through the origin} in the algebra to one-parameter subgroups in the group.
\added{For matrix groups it is given by the matrix exponential.}
Since the exponential map is not necessarily injective, it does not have a global inverse.
The \emph{Lie group logarithm},
\begin{equation}
    \log \colon U \subseteq G \to \gfrak,
\end{equation}
is defined as the diffeomorphism inverse of $\exp$, restricted to a sufficiently small neighborhood $U$ of the identity element $e$.
Distances can be defined via the Lie group logarithm,
\begin{equation}\label{eq:lie_distance}
    d_G(g, h) = \norm{\log\bigl(g^{-1} h\bigr)}_{\gfrak},
\end{equation}
where $\norm{\cdot}_{\gfrak}$ denotes a suitable weighted norm on the Lie algebra to handle unit discrepancies. \added{Concrete expressions for these operations on $SE(3)$ are provided in Sec.~\ref{ssec:pose_ds}.}

%% file: sections/4_methods.tex
\section{Curve-induced Dynamical systems on Riemannian manifolds and Lie groups}\label{sec:dsmp_methods}

Let $M$ be a smooth, finite-dimensional,
connected, and complete
Riemannian manifold.
Given reference data, we want to fit a smooth non-self-intersecting curve $\gamma \colon [0, 1] \to M$ on the data,
and construct an \emph{autonomous continuous-time \ac{ds}} of the form,
\begin{equation}
    \dot{x} = f(x),
\end{equation}
with the equilibrium point $x^{\ast} = \gamma(1)$, such that $f(x^{\ast}) = 0$, where $0$ denotes the zero vector of appropriate dimension.
The objective is to guide the system state $x(t) \in M$ to converge to $\gamma$ and propagate along it toward $x^{\ast}$.
We want $f$ to be sufficiently smooth such that for any initial condition $x(0) = x_0 \in \mathcal{D} \subseteq M$, a unique solution $x(t)$ exists for all $t \geq 0$, 
where the domain $\mathcal{D}$ depends on the global geometry of the pair $(M, \gamma)$. 

Where appropriate, we point out differences when dealing with a Lie group $G$ instead of a Riemannian manifold $M$. If not specified otherwise, one may simply replace $M$ by $G$, and $d_M$ by $d_G$, as well as apply group operations accordingly. Further, instead of tangent vectors one would then deal with Lie algebra elements.

In the following subsections we first explain how such a curve $\gamma$ can be obtained from reference data in Sec.~\ref{ssec:curve_fitting}.
Then, we discuss the construction of a DS, $f(x)$, based on this curve or any given curve that is smooth and non-self-intersecting in Sec.~\ref{ssec:DS_construction}, provide an analysis of the stability of the \ac{ds}, and explain the phase modulation layer.
In Sec.~\ref{sec:robot_manifolds}, \added{we introduce the proposed coupling between a \ac{ds} on $SE(3)$ and a curve of SPD-matrices defining the affine transformation of the velocity vectors}.

\subsection{Constructing Curves on Manifolds}\label{ssec:curve_fitting}

Given reference data, such as user-demonstrated trajectories or data obtained from an optimal control problem, we can fit a curve $\gamma$ defined by a set of parameters $\Theta$, \ie $\gamma(\added{\tau}; \Theta)$. Without loss of generality, we may assume the phase-variable $\added{\tau}$ to be in the domain $\added{\tau} \in [0, 1]$, as we can scale inputs to $\gamma$ and \added{compute} curve derivatives accordingly.
More formally, consider a set of
$K \in \N^+$ individual data sequences,
$X^{\text{ref}} = \bigcup_{k=1}^K \bigl\{ \bigl(k, \added{\tau}_i^k, x_i^k\bigr) \added{\mid i=1,\ldots,I_k} \bigr\}$, where $I_k$ denotes the number of samples in the $k$-th data sequence.
The parameters $\Theta$ of $\gamma$ can then be found by minimizing the distance on the manifold between the curve and the data-points:
\begin{alignat}{2}\label{eq:curve_fitting}
& \Theta^{*} = \!\argmin_{\Theta}  &\quad& {\sum_{k=1}^{K} \sum_{i=1}^{I_k} d_M\bigl(\gamma\bigl(\added{\tau}_i^k; \Theta\bigr), x_i^k\bigr)^2}.
\end{alignat}

As a concrete realization of curves on manifolds, we consider composite quadratic \Bezier curves, building on the Euclidean formulation in~\cite{Li25arXiv2}. To avoid distortion, the curves must fit the reference data directly on the manifold rather than in a tangent space. Tangent spaces are nonetheless used to encode individual quadratic segments, where construction is straightforward~\cite{wattersonTrajectoryOptimizationManifolds2020}, while curve evaluation, continuity constraints, and the distances in~\eqref{eq:curve_fitting} are defined on the manifold.

\begin{figure}[tb]%
    \centering%
    \begin{subfigure}[b]{0.48255\columnwidth}%
        \centering%
        \includegraphics[width=\textwidth,trim={4mm 21mm 4mm 4mm},clip]{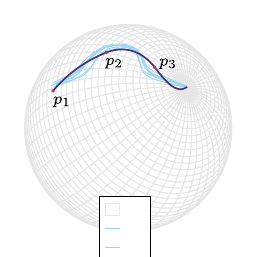}%
    \end{subfigure}%
    \hfill%
    \begin{subfigure}[b]{0.48255\columnwidth}%
        \centering%
        \includegraphics[width=\textwidth,trim={0 16mm 0 3.5mm},clip]{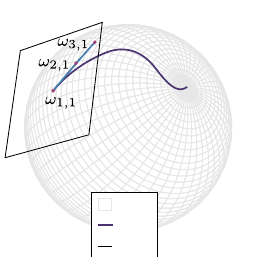}%
    \end{subfigure}%
    \\%
    \vspace{2mm}%
    \begin{subfigure}[b]{0.48255\columnwidth}%
        \centering%
        \includegraphics[width=\textwidth,trim={0 21mm 0 0.5mm},clip]{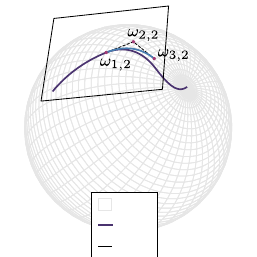}%
    \end{subfigure}%
    \hfill%
    \begin{subfigure}[b]{0.48255\columnwidth}%
        \centering%
        \includegraphics[width=\textwidth,trim={0 21mm 0 1mm},clip]{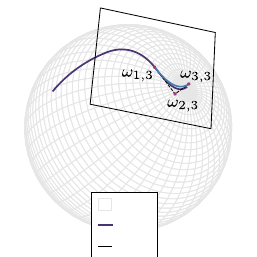}%
    \end{subfigure}%
    \caption{Curve on $S^2$ with an illustration of the curve segment definitions and control points in multiple tangent spaces. The top left shows the segment base points on the manifold, as well as the reference data in light blue. \added{The three remaining plots show the control points defined in the tangent space of the respective segment's base point.} Here, the curve is encoded with only 3 segments for better visualization of the procedure.}%
    \label{fig:tangentSpaces}%
\end{figure}

\added{We divide the complete curve into $J$ quadratic segments and parameterize it by a global phase $\tau \in [0,1]$. Let $T_j$ denote the global phase at the boundary between segments $j$ and $j+1$, with $T_0=0$ and $T_J=1$. For a given segment $j$, the corresponding global phase interval is $\tau \in [T_{j-1},T_j]$. We map this interval to a local phase $\tau_j \in [0,1]$ according to
\begin{equation} \label{eq: tau_local_global}
    \tau_j(\tau) = \left(\tau-T_{j-1}\right)/\left(T_j-T_{j-1}\right).
\end{equation}}
\added{Each segment is represented as a quadratic \Bezier curve in the tangent space at a base point $p_j \in M$. Specifically, for $j \in \{1,\ldots,J\}$,}
\begin{equation}
    u_j(\added{\tau}_j) = (1-\added{\tau}_j)^2 w_{1, j} + 2(1-\added{\tau}_j)\added{\tau}_j w_{2, j} + \added{\tau}^2_j w_{3, j}.
\end{equation}
The control points $w_{i,j}, \forall i\in\{1,2,3\}$, are expressed in this tangent space $T_{p_j}M$\added{. As the basis for a segment's tangent space, we use the first control point of the respective segment, such that} $w_{1,j}=0$ \added{is} at \added{the} origin, see Fig.~\ref{fig:tangentSpaces}.
The curve\added{-segment} $u_j(\added{\tau_j})$, defined in the tangent space, is then mapped back to a curve $\gamma_j(\added{\tau_j})$ on the manifold,
\begin{equation} \label{eq: gamma_j_curve}
    \gamma_j(\added{\tau_j}) = \exp_{p_j}(u_{j}(\added{\tau_j})).
\end{equation}
\indent To achieve $C^0$ continuity, the last control point of segment~$j$ must correspond to the same point on the manifold as the first control point of the next segment,
\begin{equation}\label{eq:c0_constraint_quad_bezier}
    p_{j+1} = \exp_{p_j}(w_{3,j}),
\end{equation}
where we note that $\exp_{p_{j+1}}(w_{1,j+1})=\exp_{p_{j+1}}(0)=p_{j+1}$,

For $C^1$ continuity~\cite{leeIntroductionSmoothManifolds2013}, the vectors $v_{2,j}$ and $v_{1,j+1}$ lying in different tangent spaces, $T_{p_j}M$ and $T_{p_{j+1}}M$, respectively, with $v_{i,j}=w_{i+1,j}-w_{i,j}$ being the direction vector between two consecutive control points, must satisfy the constraint:
\begin{equation}\label{eq:c1_constraint_quad_bezier}
    \added{v_{1,j+1} = d(\exp_{p_j})_{w_{3,j}}\bigl(v_{2,j}\bigr).}
\end{equation}
\added{This follows from differentiating~\eqref{eq: gamma_j_curve} by the chain rule: the segment velocity is $\dot{\gamma}_j(\tau_j) = d(\exp_{p_j})_{u_j(\tau_j)}\bigl(\dot{u}_j(\tau_j)\bigr)$, where $d(\exp_{p_j})_{u} \colon T_{p_j}M \to T_{\exp_{p_j}(u)}M$ is the differential of the exponential map, and on the outgoing side $u_{j+1}(0) = 0$ with $d(\exp_{p_{j+1}})_{0} = \mathrm{id}$. Segments of equal phase duration are assumed.}

\added{For Lie groups, the control points $w_{i, j}$ can be expressed in the common
Lie algebra $\gfrak \simeq T_eG$. The manifold-valued segment is then evaluated as $\gamma_j(\tau_j) = p_j\exp(u_j(\tau_j))$. Consequently, with $\exp_{p_j}(\cdot)$ becoming $p_j\exp(\cdot)$, $C^0$ continuity is enforced by $p_{j+1} = p_j\exp(w_{3,j})$ and $C^1$ continuity again by~\eqref{eq:c1_constraint_quad_bezier}.}

The final composite curve $\gamma$ is composed of the set of all the individual curves $\{\gamma_j(\added{\tau_j})\}_{j=1}^J$, with their phase \added{as in~\eqref{eq: tau_local_global}}:
\begin{equation}\label{eq:curve_gamma}
    \added{\gamma(\tau) = \gamma_j(\tau_j(\tau)),
    \qquad
    \tau \in [0,1],\quad
    j\in\{1,2,\ldots,J\}.}
\end{equation}
\added{With $\Theta$ denoting the initial base point $p_0$ and the control points $w_{i,j}$ of all segments, where $i\in\{1,2,3\}$ and $j\in\{1,\ldots,J\}$, we solve the curve-fitting problem~\eqref{eq:curve_fitting} by optimizing
these variables to minimize the discrepancy between the resulting manifold-valued curve and the reference data. The optimization is subject to the continuity constraints~\eqref{eq:c0_constraint_quad_bezier} and~\eqref{eq:c1_constraint_quad_bezier}, which enforce positional and tangent continuity between consecutive segments. The solution consequently defines the complete smooth curve (Fig.~\ref{fig:tangentSpaces}).}

\subsection{Dynamical System Construction}\label{ssec:DS_construction}

Equipped with a curve $\gamma$, we want to define a \added{\ac{ds}} that balances attraction toward the curve and continuation along the curve direction. For this, we require the closest point on the curve given a query point on the manifold. 

\begin{definition}[Closest Point Projection]\label{def:closest_point_projection}
    For any state $x \in M$, we define the projection map $\proj \colon M \to \gamma$ as the point on the curve $\gamma$ minimizing the geodesic distance:
    \begin{equation} \label{eq:projection}
        \proj(x) = \gamma(\tilde{\added{\tau}}(x)), \ \text{where} \ \tilde{\added{\tau}}(x) = \argmin_{\added{\tau} \in [0, 1]} d_{M}(x, \gamma(\added{\tau})).
    \end{equation}
\end{definition}

This map is well-defined and smooth on the domain $\mathcal{D} = M \setminus \mathcal{C}_{\text{sing}}$, where $\mathcal{C}_{\text{sing}}$ denotes the \emph{cut locus} of the curve $\gamma$.
$\mathcal{C}_{\text{sing}}$ is the set of all points $x \in M$ where the distance minimizer to $\gamma$ is not unique, or, in other words, it represents the set of points where the closest-point projection is ill-posed. This set encapsulates singularities arising from both the local curvature of the embedding (\eg, focal points) and the global topology of the manifold (\eg, antipodal sets). 
Whenever a projection is not unique, we choose the largest phase $\tilde{\added{\tau}}$ out of all points on the curve $\gamma(\added{\tau})$ with minimal distance from the query point~$x$ \added{(App.~\ref{ssec: part3_convergence_local_max})}.

\begin{figure}[bt]%
\centering%
\begin{subfigure}{0.467\columnwidth}%
    \centering%
    \includegraphics[width=\linewidth]{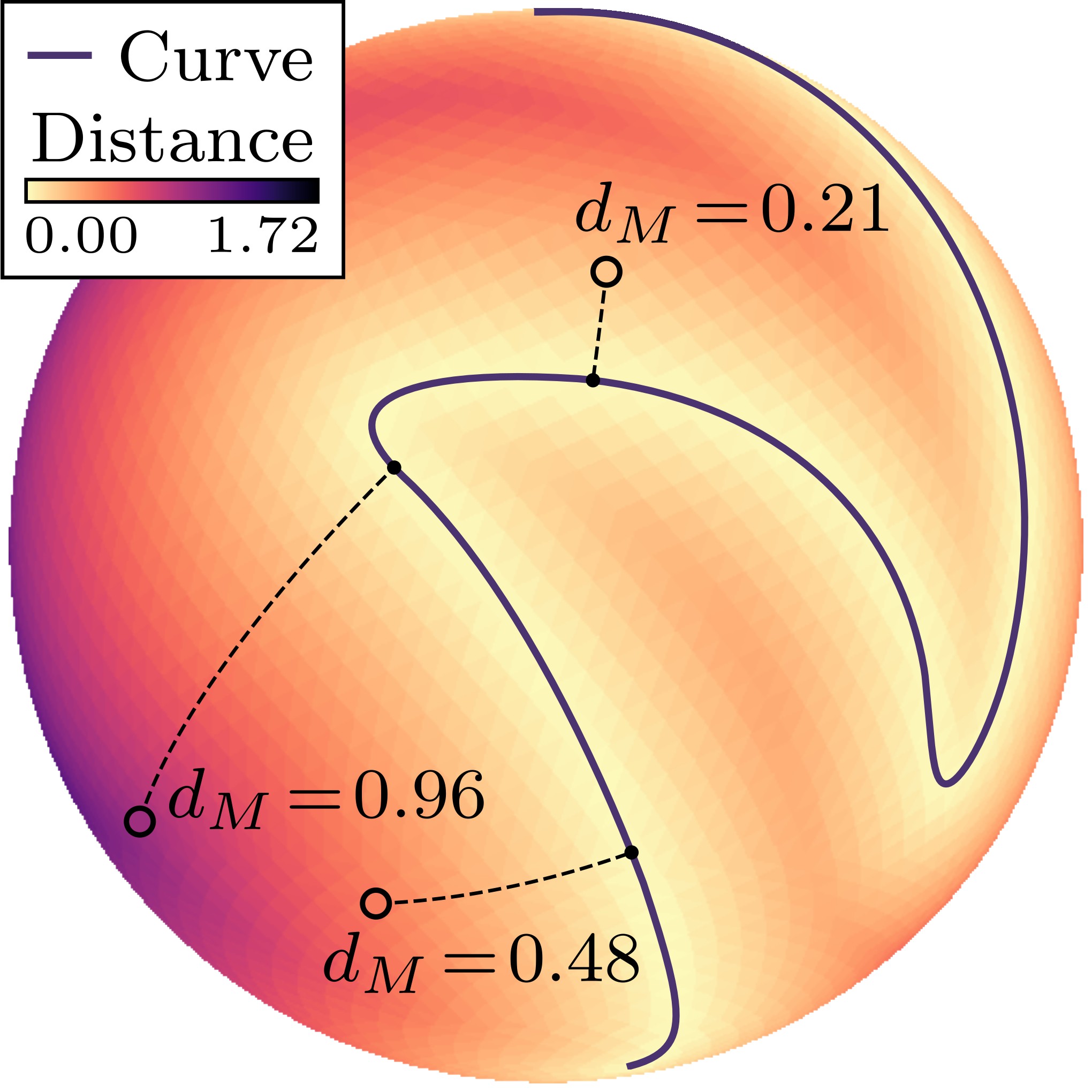}%
    \caption{}%
    \label{fig:distance_field_s2}%
\end{subfigure}%
\hfill%
\begin{subfigure}{0.467\columnwidth}%
    \centering%
    \includegraphics[width=\linewidth]{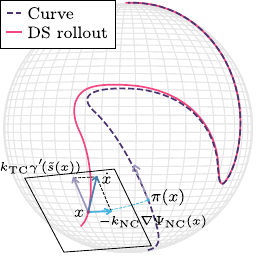}%
    \caption{}%
    \label{fig:s2_illustration_trajectory}%
\end{subfigure}%
\caption{
(a) Distance field for a curve on $S^2$ with examples of projections and geodesic distances on the manifold. (b) A trajectory of the \ac{ds} converging toward the curve with the tangential (TC) and normal (NC) term of the \ac{ds} in the tangent space at the query point.
The velocities are scaled for the illustration.
}
\label{fig:ds_illustration_s2}%
\end{figure}

Since the optimization problem encapsulated by $\proj$ in~\eqref{eq:projection} is on a one-parameter curve, it is straightforward and efficient to solve.
Fig.~\ref{fig:distance_field_s2} shows a curve and its distance field along with some closest point projections on the $S^2$-manifold.
Note that this distance \emph{field} is purely illustrative as our method only requires to perform the closest point projection for the current query point.

Given a curve $\gamma$ and the projection operator $\proj$ from Definition~\ref{def:closest_point_projection}, we define a DS as the superposition of a  term \emph{tangential} to $\gamma$ (TC; to propagate along the curve) and a term \emph{normal} to $\gamma$ (NC; to eliminate cross-track error).

\begin{definition}[Dynamical System] \label{def:dynamical_system}
    \label{def:ds}
    For all $x \in \added{M}$,
    the closed-loop DS is governed by:
    \begin{equation}  \label{eq:ds}
        \dot{x} = f(x) \coloneqq \underbrace{k_{\text{TC}} \PT{\proj(x)}{x}(\gamma'(\tilde{\added{\tau}}(x)))}_{\text{Propagation}} \, \underbrace{-k_{\text{NC}} \nabla \Psi_{\text{NC}}(x)\vphantom{\PT{\proj(x)}{x}}}_{\text{Convergence}}, 
    \end{equation}
    where
    $k_{\text{TC}}, k_{\text{NC}} > 0$ are positive scalar gains,
    $\pi(x) = \gamma(\tilde{\added{\tau}})$ is the closest point on the curve to $x$ according to~\eqref{eq:projection},
    $\gamma'(\tilde{\added{\tau}}) \in T_{\proj(x)}M$ is the velocity of the curve at $\pi(x)$ pointing along the curve, and $\PT{\proj(x)}{x}$
    transports $\gamma'(\tilde{\added{\tau}})$ along the geodesic from $\proj(x)$ to $x$.
    The normal energy potential is
    \begin{equation}\label{eq:normal_energy_potential} 
    \Psi_{\text{NC}}(x) = \frac{1}{2} d_M^2(x, \proj(x)).    
    \end{equation}
    Its negative gradient, $-\nabla \Psi_\text{NC}(x)$, corresponds to the tangent vector at $x$ of the geodesic connecting $x$ to $\proj(x)$, 
    and points toward the curve.
    
\end{definition}
The propagation term represents the tangential desired velocity $\gamma'(\tilde{\added{\tau}})$ at the current point $x$, while the convergence term drives the system toward the curve.
By the First Variation of Arc Length (Gauss' Lemma)~\added{\cite{leeIntroductionRiemannianManifolds2018}}, these two vectors are orthogonal at $x$ where $\tilde{\added{\tau}}(x) \in (0, 1)$, which ensures that the propagation term does not degrade the convergence to the curve. 
For Lie groups, the propagation term, without $k_{\text{TC}}$, corresponds to the Lie algebra element \added{obtained by left translation to the identity},
\added{$d(L_{\gamma(\tilde{\tau})^{-1}})_{\gamma(\tilde{\tau})}\bigl(\gamma'(\tilde{\tau})\bigr)$, which for matrix groups is $\gamma(\tilde{\tau})^{-1} \gamma'(\tilde{\tau})$}.
The negative gradient of the normal energy potential corresponds to the Lie algebra element $\log(x^{-1} \pi(x))$.

Note a crucial difference to the approach in Euclidean space as described in~\cite{Li25arXiv2}. In Euclidean space, the propagating velocity vector on the curve, $\gamma'(\tilde{\added{\tau}})$, can be directly applied to $x$ as the tangential velocity vector.
In our case of Riemannian manifolds, $\gamma'(\tilde{\added{\tau}})$ is initially computed in the tangent space at $\proj(x)$, while the final command needs to be expressed in the tangent space at $x$, \ie $\PT{\proj(x)}{x}(\gamma'(\tilde{\added{\tau}}))$. 
\added{In addition, in Euclidean space, the closest point can be obtained analytically for each segment by solving a cubic polynomial~\cite{Li25arXiv2}, whereas on a Riemannian manifold or Lie group, the closest point on a curve generally has to be determined numerically.}

\subsubsection{Behavior Toward Goal} \label{ssec: behavior_to_goal}

In order to ensure that the DS \added{smoothly} comes to a stop at the final point on the curve, $\gamma(1)$, independently of the provided reference data, we scale the tangential vector
to zero as we approach the goal.
This is done by defining the \added{curve derivative $\gamma'$} in~\eqref{eq:ds} as 
\begin{equation}\label{eq:twist_scaling_to_zero}
\added{\gamma'(\tilde{\added{\tau}}) = \zeta(\tilde{\added{\tau}}) (\gamma^{\text{c}})'(\tilde{\tau}), \ \text{where} \ \zeta(\tilde{\added{\tau}}) = \left(1 - \tilde{\added{\tau}}^{k_{\text{g}}}\right),}
\end{equation}
with \added{original curve derivative $(\gamma^{\text{c}})'(\tilde{\tau})$}, and configuration parameter $k_{\text{g}} > 1$. A smaller value for $k_{\text{g}}$ means that the scaling to zero becomes more significant further away from the goal.
Since $\zeta$ is smooth with $0 \leq \zeta \leq 1$, 
it only achieves a phase reparameterization and does not affect the system's stability.

\subsubsection{\added{Cyclic motions}}  \label{sssec: cyclic_motions}
\added{
Certain motions, such as repeated exercises in rehabilitation~\cite{schedler2020effects}, are inherently cyclic and can therefore be represented by repeatedly traversing a single demonstration. To enable such cyclic motions, we remove the phase-dependent scaling at the final point of the curve in~\eqref{eq:twist_scaling_to_zero}, i.e. $\zeta = 1$. We additionally impose closure of the curve by requiring the final point to coincide with the initial point and enforcing continuity of the velocity at the junction:
\begin{subequations}
\begin{align}
    \exp_{p_1}(w_{1,1}) &= \exp_{p_J}(w_{3,J}),\\
    v_{1,1} &= d(\exp_{p_J})_{w_{3,J}}\bigl(v_{2,J}\bigr).
\end{align}
\end{subequations}
Here, the differential of the exponential map maps the velocity at the end of the last segment into the tangent space at the junction point, as in~\eqref{eq:c1_constraint_quad_bezier}, ensuring that the velocity flows continuously from the final segment into the first segment.}

\subsubsection{Generic Riemannian Manifolds}\label{ssec:note_on_general_RM}
\begin{figure}[tb]
\centering
\begin{subfigure}[b]{0.365\columnwidth}%
    \centering
    \includegraphics[width=\linewidth]{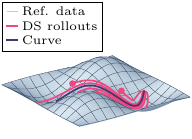}%
\end{subfigure}\hfill
\begin{subfigure}[b]{0.359\columnwidth}%
    \centering
    \includegraphics[width=\linewidth]{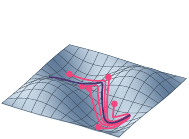}%
\end{subfigure}\hfill
\begin{subfigure}[b]{0.264\columnwidth}%
    \centering
    \includegraphics[width=\linewidth]{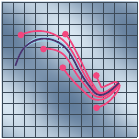}%
\end{subfigure}
\caption{\added{%
\ac{dsmpman} on a curved surface given implicitly, where none of the Riemannian operations have a closed form and all are obtained by integrating the geodesic equation. The curve is fitted to the demonstrations and the induced \ac{ds} converges to it and propagates to the goal. Depicted are different views of the same \ac{ds} and manifold.}}
\label{fig:levelset}
\end{figure}

\phantom{}\added{The examples and use-cases in our work focus on Riemannian manifolds and Lie groups, which encompass a rich variety of geometric structures relevant to robotics, including $S^2$, $SO(3)$, $SE(3)$, and $\SPDn$~\cite{Calinon26AR}. In Definition~\ref{def:closest_point_projection} and \ref{def:dynamical_system}, we assume that distances on such manifolds or Lie groups, as well as transport operations, can be computed analytically or numerically in real time. For the commonly used manifolds listed above, closed-form expressions are available, while more general manifolds may require numerical integration or optimization. Using JAX~\cite{jax2018github, jaxopt_implicit_diff}, such computations can often be performed efficiently enough for near real-time operation.
Such manifolds arise in robotics whenever the geometry is induced rather than given. For example, a task is constrained to a curved workpiece surface described by a level set, or a configuration space equipped with the kinetic-energy metric, in which velocity commands are weighted by the robot's inertia matrix and the geodesics are its unforced motions.

Although a systematic evaluation on more general manifolds is beyond the scope of this paper, we illustrate the extension to a generic Riemannian manifold using an implicit surface defined as the zero level set of a scalar function. Its geodesics and associated operations have no closed-form expressions: the exponential map is obtained by integrating the geodesic equation, the logarithm by shooting for the initial velocity reaching the target, the distance from the geodesic length, and parallel transport by integration along the geodesic. The closest point is found by selecting nearby curve samples in the embedding space and evaluating geodesic distances only for these candidates, since shooting is reliable only locally. 
Fig.~\ref{fig:levelset} demonstrates the \ac{ds} and curve fit of \ac{dsmpman} on this surface, where a single query of the resulting \ac{ds} takes \SI{0.3}{\milli\second}. 
}

\subsection{Stability Analysis} \label{sec:stab_analysis}
\begin{figure*}[bt]%
\centering%
\begin{subfigure}{0.22\textwidth}%
    \centering%
    \includegraphics[width=\linewidth]{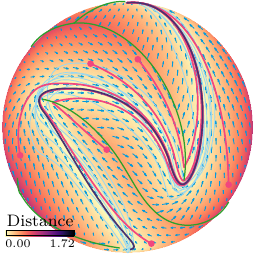}%
    \caption{}%
    \label{fig:lasa_nshape_ridges_a}%
\end{subfigure}%
\hfill%
\begin{subfigure}{0.22\textwidth}%
    \centering%
    \includegraphics[width=\linewidth]{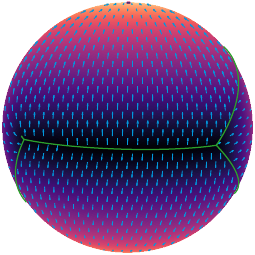}%
    \caption*{\vphantom{(a)}}
    \label{fig:lasa_nshape_ridges_b}%
\end{subfigure}%
\hfill%
\begin{subfigure}{0.22\textwidth}%
    \centering%
    \includegraphics[width=\linewidth]{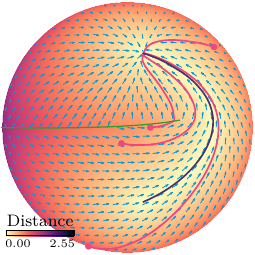}%
    \caption{}%
    \label{fig:lasa_cshape_ridges_a}%
\end{subfigure}%
\hfill%
\begin{subfigure}{0.22\textwidth}%
    \centering%
    \includegraphics[width=\linewidth]{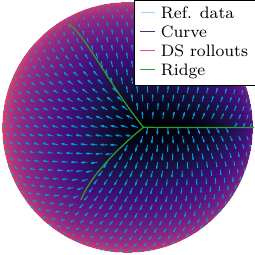}%
    \caption*{\vphantom{(a)}}
    \label{fig:lasa_cshape_ridges_b}%
\end{subfigure}%
\caption{%
\added{(a) LASA N-Shape on $S^2$, front and back views. (b) C-Shape on $S^2$, front and back views. The singular set is indicated in green and vectors represent the velocity induced by the \ac{ds}.}}
\label{fig:lasa_ridges}%
\end{figure*}

\added{Formally establishing global asymptotic convergence of \ac{ds}s on Riemannian manifolds remains challenging due to topological limitations and the possible presence of undesired equilibria or singularities.}
For example, the Lie group $SO(3)$ is double covered by unit quaternions, meaning that two distinct quaternions, $q$ and $-q$, represent the same rotation in $SO(3)$\added{. This can lead to the \textit{unwinding problem}, where the system makes a full rotation instead of taking the shortest path.
Global asymptotic stabilization on $SO(3)$ is not possible using classical continuous state feedback, although hybrid feedback can address this issue by switching between suitably constructed feedback laws~\cite{mayhew2013synergistic}.
}
In addition, the curve $\gamma(\added{\tilde{\tau}})$ will likely cause singular points or singular regions in the \ac{ds}~\cite{leeIntroductionSmoothManifolds2013}, and trajectories could exist that converge to the singular set instead of the curve (see the example in Fig. 5.6 in~\cite{yaoGuidingVectorFields2023}).

\added{When observing Fig.~\ref{fig:lasa_ridges}, for a 1D curve on $S^2$, the singular set $\mathcal{C}_{\text{sing}}$ contains points where the closest point projection onto the curve is non-unique.}
Geometrically, $\mathcal{C}_{\text{sing}}$ forms the ``ridges'' of the distance potential $\Psi_\text{NC}$.
In directions transversal to these ridges, the distance potential is concave, \ie the ridges are local maxima and represent unstable equilibria of $\Psi_\text{NC}$~\cite{leeGeometricTrackingControl2010}. 
\added{For the stability analysis, we consider trajectories that \textit{start} in the domain $\mathcal{D} = M \setminus \mathcal{C}_{\text{sing}}$, which is realistic as $\mathcal{C}_{\text{sing}}$ has measure zero for a 1D curve on an $n$-dimensional manifold.
In addition, as the ridges represent unstable equilibria, the assumption that no trajectories exist that converge to $\mathcal{C}_{\text{sing}}$ is valid. 
However, as observed in Fig.~\ref{fig:lasa_ridges}, trajectories can exist that \textit{pass through} $\mathcal{C}_{\text{sing}}$. }

\added{The analysis of the stability of the \ac{ds} will therefore proceed in three parts. 
Firstly, we analyze local convergence to the curve, $\gamma(\tau(x))$, which takes inspiration from~\cite{lee2025behavior}. Secondly, we will extend upon this by establishing local convergence to the goal, $\gamma(1)$. 
For these first two parts, we assume to operate in the domain $\mathcal{D} = M \setminus \mathcal{C}_{\text{sing}}$, and that no trajectories exist that start in $\mathcal{C}_{\text{sing}}$, converge to $\mathcal{C}_{\text{sing}}$, or pass through $\mathcal{C}_{\text{sing}}$. This means that on $\mathcal{D}$, the minimizer of the distance, $d^2_M\bigl(\gamma\bigr)$, is unique.
The local convergence analysis further requires the relevant sublevel sets of the Lyapunov function to be compact.
Lastly, an analysis is provided for trajectories that pass through $\mathcal{C}_{\text{sing}}$ for a more global notion of convergence.}

\subsubsection{\added{Local convergence to the curve}}

\added{To analyze local convergence to the curve on $\mathcal{D}$, consider the Lyapunov function given by the squared minimum geodesic distance to the trajectory,
\begin{equation} \label{eq: lyapunov_curve_main}
V(x)=\Psi_{\mathrm{NC}}(x)
=\frac{1}{2}d_M^2(\gamma,x),
\end{equation}For $\tilde{\tau}\in(0,1)$, the geodesic tangent and curve tangent are orthogonal by the first variation of Arc Length~\cite{leeIntroductionRiemannianManifolds2018}. At $\tilde{\tau}=1$, the curve velocity is zero by construction using~\eqref{eq:twist_scaling_to_zero}, while at $\tilde{\tau}=0$ the endpoint contribution is non-positive. Consequently,
\begin{equation} \label{eq: vdot_curve_main}
\dot V(x)
\leq
-k_{\mathrm{NC}}d_M^2(\gamma,x)
\leq 0.
\end{equation}
with strict decrease for $x\notin\gamma$. Thus, the \ac{ds} locally converges to the curve $\gamma(\tau)$ on $\mathcal{D}$.}
\added{The detailed derivation of $\dot V$, including the endpoint cases $\tilde{\tau}=0$ and $\tilde{\tau}=1$, is provided in App.~\ref{ssec: part1_local_to_curve}.}

\subsubsection{\added{Local convergence to the goal}}

\added{To establish convergence to the goal $x^{*}=\gamma(1)$, we augment the Lyapunov function in~\eqref{eq: lyapunov_curve_main} with the tangential potential
$\Psi_{\mathrm{TC}}(x)=\frac{1}{2}\left(1-\tilde{\tau}(x)\right)^2,$
\begin{equation}\label{eq:lyapunov_full_main}
\begin{aligned}
V(x)=\Psi_{\mathrm{NC}}(x)+\Psi_{\mathrm{TC}}(x).
\end{aligned}
\end{equation}
Leveraging the result of~\eqref{eq: vdot_curve_main} and again analyzing the cases where $\tilde{\tau} = 0$, $\tilde{\tau} \in (0, 1)$, and $\tilde{\tau} = 1$ separately, we obtain the derivative, as explored in App.~\ref{ssec: part2_local_conv_goal},
{\small
\begin{subequations}
\begin{align}
\added{\dot{V}(x)}
&\added{= - k_{\text{NC}} d_{M}^2(\gamma, x)
- \frac{k_{\text{TC}}}{\beta(x)} (1 \scalebox{0.8}[1]{$-$} \tilde{\tau}) C \leq 0,}
&& \added{\forall \tilde{\tau} \in (0, 1],} 
\label{eq:vdot_goal_main_2}\\
\added{\dot{V}(x)}
&\added{\leq - k_{\text{NC}} d_M^2 < 0,}
&& \added{\text{if } \tilde{\tau} = 0,}
\end{align}
\end{subequations}}%
with strict inequality of~\eqref{eq:vdot_goal_main_2} when $x \neq x^{*}$, and $\beta(x)>0$, $C>0$ for $\tilde{\tau} < 1$, and $C = 0$ for $\tilde{\tau} = 1$.
Therefore, all trajectories that start in $\mathcal{D}$ and stay in $\mathcal{D}$, converge to the equilibrium point $x^{*} = \gamma(1)$.}

\subsubsection{\added{Passing through $\mathcal{C}_{\text{sing}}$}}
\added{Trajectories may encounter the singular set $\mathcal{C}_{\mathrm{sing}}$. As before, we assume that trajectories do not originate from or converge to $\mathcal{C}_{\mathrm{sing}}$, and will focus on analyzing trajectories that \textit{pass through} singular points which have multiple (distinct) points on the curve with minimal distance, denoted by $\mathcal{C}_{\mathrm{sing, 1}}$.
Following the analysis in App.~\ref{ssec: part3_convergence_local_max}, among the phase values of all closest-point projections to the curve, captured in $\mathcal{P}_{\tau}$, the maximum phase value $\tilde{\tau}^{*} \in \mathcal{P}_{\tau}$ is selected, and used for defining the \ac{ds}. 
With this choice of $\tilde{\tau}^{*}$, we establish that the energy strictly decreases within regular regions and is non-increasing at crossings of $\mathcal{C}_{\mathrm{sing, 1}}$, and therefore trajectories starting in $\mathcal{D}$ converge to $
x^{*}=\gamma(1)$.
This result requires a finite number of crossings of $\mathcal{C}_{\text{sing,1}}$. The detailed singular-set analysis is provided in App.~\ref{ssec: part3_convergence_local_max}.}

\subsection{Phase Modulation}\label{sec:phase_modulation}
\added{Via the phase modulation layer, the \ac{dsmpman} framework can decouple the construction of the phase profile from that of the spatial profile.}
The phase and spatial profiles are parameterized using concatenated \Bezier curves on $\mathbb{R}_{\geq 0}$ and the manifold, respectively.
Adjusting the timing separately is beneficial in scenarios where, for example, the time profile should be optimized to remain within certain velocity limits, or collisions should be avoided with dynamic agents by slowing down or speeding up along the curve, while the spatial profile remains intact for user interpretability. Similar to a curve in Euclidean space, we can represent the evolution of the phase variable $\added{\tau}$ of $\gamma(\added{\tau})$ by a curve $\gamma^{\added{\tau}}(t)$, with end-time $T$ and $J$ segments:
\begin{equation} \label{eq:curve_gamma_timing}
\begin{aligned}
    & \added{\tau} = \gamma^{\added{\tau}}(t) = \Psi(t/T) \, w, \quad t \in [0, T],
\end{aligned}
\end{equation}
where $w$ is a vector concatenation of all $w_{i,j}, \forall i\in\{1,2,3\}$ and $j\in\{1,2,\ldots,J\}$. Here, $\Psi(t)$ is a matrix containing Bernstein polynomial coefficients~\cite{Li25arXiv2}.
This representation allows us to flexibly define cost functions to build a phase and spatial profile separately. 
By restricting the phase $\added{\tau}$ to be in the domain $[0, 1]$ and allowing no backward motion along the curve, \ie $\left(\gamma_{\added{\tau}}\right)'(t)>0$, the time adaptation does not affect the stability analysis in Sec.~\ref{sec:stab_analysis}.
The DS gets affected by the phase modulation layer as the propagation term in~\eqref{eq:ds} depends on the derivative of the curve. Since the spatial curve $\gamma$ is parametrized by a phase variable $\added{\tau}(t)$, the chain rule yields
\begin{equation}
    \added{\dot{\gamma}(t)} = \frac{\delta\gamma}{\delta \added{\tau}}\,\frac{\delta \added{\tau}}{\delta t} = \frac{\delta\gamma}{\delta \added{\tau}}\,(\gamma^{\added{\tau}})'(t),
\end{equation}
where $\frac{\delta \gamma}{\delta \added{\tau}}$ describes the derivative of the spatial curve with respect to the phase.

%% file: sections/4b_relevant_manifolds.tex
\section{\added{Coupling \ac{ds}s and curves on Robotics-Relevant Manifolds}}\label{sec:robot_manifolds}
\added{An integral part of our approach, is the proposed coupling between \ac{ds}s and curves that lie on different manifolds, given the fact that} the proposed \ac{dsmpman} is applicable to a broad class of Riemannian manifolds and Lie groups standard in robotics.
\added{In Sec.~\ref{ssec:pose_ds_damping}, a coupling mechanism is provided between a \ac{ds} of rigid body motions and a curve of variable affine transformation matrices for the desired velocity, which we will refer to as damping matrices.}
\added{Before that, Sec.~\ref{ssec:pose_ds} and Sec.~\ref{sec:spd} introduce the} construction of the exponential and logarithmic map, its distance and transformation for $SE(3)$ and $\SPD{n}$\added{, respectively}. Further details are available in~\cite{sola_micro_2021} for $SE(3)$ and other Lie groups, and in~\cite{pennecRiemannianFrameworkTensor2006} for $\SPD{n}$.
Examples of other Riemannian manifolds and Lie groups can be found in our code.

\subsection{Dynamical System of 6-DoF Poses}\label{ssec:pose_ds}

Of particular interest for robotics is the group of rigid body motions, $SE(3)$. 
It is double-covered by the Lie group
$G = \UQR$, where $\UQ = \{q \in \Hspace : \norm{q}=1\}$ is the group of unit quaternions, $\Hspace$ denotes the algebra of quaternions, and $\ltimes$ denotes the semidirect product. Elements $g \in G$ are pairs $(q, t)$, where $q \in \UQ$ represents rotation, and $t \in \R^3$ is a global translation vector. The Lie algebra of this group is identified with $\se$, where elements are pairs $(\omega, v) \in \R^3 \times \R^3$, where $\omega$ is the angular velocity and $v$ is the linear velocity, both expressed \emph{in the body frame}. 
The exponential map sends $(\omega, v)\in \se$ to $(q, t) \in \UQR$ and the logarithmic map does the inverse. 
The exponential map $(q,t) = \exp(\omega,v)$, where $q$ goes to identity as $\theta \to 0$, is given by
\begin{equation}
q = \cos(\theta/2) + \frac{\omega}{\theta}\sin(\theta/2),
\quad
t = V(\omega)\,v,
\end{equation}
where $V(\omega)$ is the left Jacobian of $SO(3)$,
\begin{equation}
V(\omega)
= I + \frac{1-\cos\theta}{\theta^2}[\omega]_\times
  + \frac{\theta-\sin\theta}{\theta^3}[\omega]_\times^2,
\ \  \theta=\|\omega\|.
\end{equation}
The logarithmic map $(\omega,v) = \log(q,t) $ is given by
\begin{equation}
\omega = \log(q),
\quad
v = V^{-1}(\omega)\,t,
\end{equation}
with
\begin{equation}
V^{-1}(\omega)
= I - \frac12[\omega]_\times
  + \frac{1}{\theta^2}\!\left(1-\frac{\theta\sin\theta}{2(1-\cos\theta)}\right)\![\omega]_\times^2.
\end{equation}

We equip the Lie algebra $\gfrak \cong \se$ with a weighted inner product to handle the unit inconsistency between rotation (radians) and translation (meters). We introduce a characteristic length $L_c$ and a weighting parameter $\eta = L_c^2$.
Consistent with the general definition for Lie groups, we utilize the log-Euclidean distance:
\begin{equation}
    d_G(g_1, g_2) = \norm{\log\bigl(g_1^{-1} g_2\bigr)}_{\gfrak} = \sqrt{\eta \norm{\omega}^2 + \norm{v}^2}.
\end{equation}

\subsection{Symmetric Positive Definite Matrices $\SPD{n}$}\label{sec:spd}

Let $\SPDn$ be the manifold of $n \times n$ symmetric positive definite matrices, often encoding stiffness matrices, damping matrices, or cost metrics in robotics,
\begin{equation}
    \SPDn = \bigl\{P \in \mathbb{R}^{n\times n} \mid P = P^\transpose \text{ and } P \succ 0\bigr\}.
\end{equation}
Its tangent space $T_P \SPDn$ at any point $P$, is the space of symmetric matrices $\mathcal{S}^n$.
The closed-form expressions for the exponential and logarithmic maps are:
\begin{align}
    \exp_P(V) &= P^{\frac{1}{2}} \, \mathrm{expm}\bigl(P^{-\frac{1}{2}} V P^{-\frac{1}{2}}\bigr) P^{\frac{1}{2}}, \\
    \log_P(Q) &= P^{\frac{1}{2}} \, \mathrm{logm}\bigl(P^{-\frac{1}{2}} Q P^{-\frac{1}{2}}\bigr) P^{\frac{1}{2}},
\end{align}
with $\mathrm{expm}(\cdot)$ and $\mathrm{logm}(\cdot)$ being the matrix exponential and logarithm, respectively.
The geodesic distance between two SPD matrices $P, Q \in \SPDn$ is:
\begin{align}
    d_{\SPDn}(P, Q) &= \bigl\lVert\ln\bigl(P^{-\frac{1}{2}} Q P^{-\frac{1}{2}}\bigr)\bigr\rVert_\text{F} = \!\Biggl( \sum_{i=1}^n \ln^2 \lambda_i \Biggr)^{\frac{1}{2}},
\end{align}
where $\norm{\cdot}_\text{F}$ is the Frobenius norm, and $\lambda_i$ are the real, positive eigenvalues of $P^{-1} Q$.
Parallel transport of $V \in \Sym$ from $P \in \SPDn$ to $Q \in \SPDn$ is computed as:
\begin{equation}
    \PT{P}{Q}(V) = A V A^\transpose, \quad \text{where } A = \bigl( Q P^{-1} \bigr)^{\frac{1}{2}}.
\end{equation}

\subsection{Coupling Poses in $SE(3)$ and Variable Matrices in $\SPD{6}$}\label{ssec:pose_ds_damping}

\begin{figure}[tb]
\centering
\includegraphics{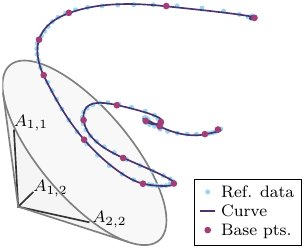}
\caption{%
Spline on $\SPD{2}$, which corresponds to the $\SPD{2}$-spline used in Fig.~\ref{fig:variable_spd_2D_rn2_vectorfield}, obtained from the covariance. The axes indicate the diagonal ($A_{1,1}$ and $A_{\added{2},2}$) and off-diagonal ($A_{1,2}$) terms of the $2\times 2$ matrices.
\added{The covariance is obtained from $K=7$ demonstrations of one LASA shape, consisting of 100 samples each.}
}
\label{fig:spd2_spline}
\end{figure}

\begin{figure}[tb]
\centering
\includegraphics[width=0.925\columnwidth]{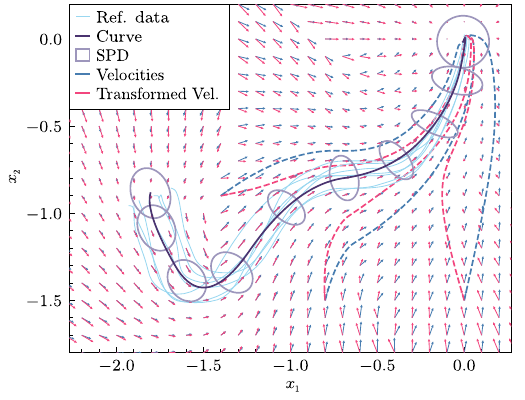}
\caption{%
Vector fields of a \added{\ac{ds}} in $\R^2$ without (blue) and with (pink) variable transformation matrices applied to the velocities, estimated from the reference data covariance. The velocity vectors not only change length but also direction, highlighting the presence of non-zero off-diagonal elements in the transformation matrices. 
\added{The ellipses correspond to the velocity gain matrices in~\eqref{eq: vel_gain_coupling} with regularization term $d=1$.}
The dashed lines are forward simulations of the unmodified and transformed velocities.}
\label{fig:variable_spd_2D_rn2_vectorfield}
\end{figure}

For the dressing task shown in Fig.~\ref{fig:dressing}, the curve $\gamma_{SE(3)}$ encodes the nominal trajectory for pulling the sleeve over the arm\added{, from the hand toward the shoulder}. Meanwhile, the \added{\ac{ds}}, $f_{SE(3)}$, extends the velocity profile to states off the curve, $x \notin \gamma_{SE(3)}$, enabling robust handling of disturbances and arbitrary initial conditions on the group. Along $f_{SE(3)}$ encoding poses, we incorporate a continuous profile of variable gain matrices, $\gamma_{\SPD{n}}$, that affect the resulting behavior. 
The required $\SPD{n}$ matrices at different poses can be provided by an expert or come from a human motor control-study~\cite{ajoudani2015reduced}.
As an accessible proxy for a $\SPD{n}$ profile, we consider the variability of the reference data, in form of covariance matrices, and invert them~\cite{saverianoLearningStableRobotic2023, kronanderLearningCompliantManipulation2014, silverioUncertaintyAwareImitationLearning2019}.
Intuitively, the larger the local variability observed in the data, the smaller the gains should be for returning to the reference under perturbations; mathematically, this is captured by defining the gain matrix as the inverse of the covariance matrix, which can be computed via eigenvalue decomposition.
Although low covariance does not always justify high values of $\gamma_{\SPD{n}}$ when near obstacles or in contact, this approach remains appropriate for free-space motions.
Given a nominal pose curve, the idea is to estimate the covariance with respect to this curve.
\added{Since the $K$ demonstrations have different durations, we divide all time stamps by the longest of the $K$ durations.
Each trajectory is evaluated at $N$ uniformly spaced phases in $[0,1]$ using geodesic interpolation, $\exp_{x_i}\!\bigl(\alpha\log_{x_i}(x_{i+1})\bigr)$. No time warping is applied. For demonstrations ending before $\tau=1$, the final configuration is held until $\tau=1$, so towards $\tau=1$ the covariance increasingly reflects the spread of the endpoints rather than of the motion.}
After aligning and resampling the reference data in time, $\hat{X}^{\text{ref}}$, consisting of $K$ demonstrations, the mean and covariance at each phase $\added{\tau}$ can be computed as,
\begin{align}
\mu_{\added{\tau}} &= \frac{1}{K} \! \!\sum_{x \in \hat{X}^{\text{ref}}(\added{\tau})} \!\! \log_{\gamma(\added{\tau})}(x),\\
\Sigma_{\added{\tau}} &= \frac{1}{\added{K\scalebox{0.8}[1]{$-$}1}}\! \! \! \! \!\sum_{x \in \hat{X}^{\text{ref}}(\added{\tau})} \!\! \bigl(\log_{\gamma(\added{\tau})}(x) - \mu_{\added{\tau}}\bigr) \bigl(\log_{\gamma(\added{\tau})}(x) - \mu_{\added{\tau}}\bigr)^\transpose.
\end{align}
We then compute the eigenvalue decomposition as $\Sigma_{\added{\tau}} = \added{V \Lambda V^\transpose}$, where $V \in \R^{n\times n}$ is an orthonormal matrix of eigenvectors of $\Sigma_{\added{\tau}}$, and $\Lambda = \mathrm{diag}(\lambda_1, \,\ldots,\, \lambda_n) \in \R^{n\times n}$ is a real diagonal matrix of eigenvalues.
While~\cite{saverianoLearningStableRobotic2023} encode all orientations in the tangent space at the goal quaternion, we estimate the covariance matrices in the nominal curve's moving tangent space and work with
full $6 \times 6$ covariance matrices.
Finally, we obtain variable velocity gain matrices, also referred to as damping matrices, as:
\begin{equation} \label{eq: vel_gain_coupling}
D_{\added{\tau}} = V \, \mathrm{diag}\biggl(\frac{1}{\lambda_1 + d}, \,\ldots,\, \frac{1}{\lambda_n + d}\biggl) V^\transpose,
\end{equation}
with \added{regularization term} $d$.
This provides us with a set of phase-indexed damping matrices, $\{D_{\added{\tau}}\}$,
to which the curve $\gamma_{\SPD{n}}$ can be fitted as in Sec.~\added{\ref{ssec:curve_fitting}}, and synchronized to the \added{\ac{ds}}, $f_{SE(3)}$, by phase.

\begin{table*}[!htb]
    \caption{Quantitative results on the LASA dataset mapped to the $S^2$-manifold\added{: mean $\pm$ standard deviation over the 26 shapes; distance metrics are geodesic distances on the unit sphere, in radians, and times are in seconds except $t_\mathrm{query}$, which is in milliseconds}.}
    \label{tab:lasas2}
    \centering
    \footnotesize
    \setlength{\tabcolsep}{4pt}
    \begin{tabular}{l*{7}{r}}
    \toprule
    & \multicolumn{1}{c}{\added{RMSE}} & \multicolumn{1}{c}{\added{DTWD}} & \multicolumn{1}{c}{\added{Path Distance}} & \multicolumn{1}{c}{Goal Distance} & \multicolumn{1}{c}{\added{$t_\mathrm{rollout}$}} & \multicolumn{1}{c}{\added{$t_\mathrm{query}$}} & \multicolumn{1}{c}{\added{$t_\mathrm{fit}$}} \\
    \midrule
    \ac{dsmpman} (ours)             & \added{0.185 $\pm$ 0.111} & \added{0.068 $\pm$ 0.046} & \added{\textbf{0.031} $\pm$ 0.021} & \added{\textbf{0.0012} $\pm$ 0.0008} & \added{0.0017 $\pm$ 0.0000} & \added{0.026 $\pm$ 0.001} & \added{\phantom{00}0.53 $\pm$ 0.02} \\
    \added{\ac{dsmpman}, single chart} & \added{0.186 $\pm$ 0.145} & \added{0.072 $\pm$ 0.068} & \added{0.054 $\pm$ 0.074} & \added{0.0023 $\pm$ 0.0013} & \added{\textbf{0.0008} $\pm$ 0.0000} & \added{\textbf{0.016} $\pm$ 0.001} & \added{\phantom{00}0.11 $\pm$ 0.00} \\
    LieFlows                         & \added{0.170 $\pm$ 0.110} & \added{\textbf{0.052} $\pm$ 0.034} & \added{0.107 $\pm$ 0.141} & \added{0.2020 $\pm$ 0.3233} & \added{0.1642 $\pm$ 0.0021} & \added{0.954 $\pm$ 0.014} & \added{983.07 $\pm$ 2.32} \\
    PUMA                             & \added{\textbf{0.152} $\pm$ 0.129} & \added{0.054 $\pm$ 0.085} & \added{0.122 $\pm$ 0.109} & \added{0.0133 $\pm$ 0.0132} & \added{0.0187 $\pm$ 0.0005} & \added{0.072 $\pm$ 0.003} & \added{343.49 $\pm$ 3.64} \\
    \added{BCSDM-GVF}            & \added{0.278 $\pm$ 0.199} & \added{0.080 $\pm$ 0.079} & \added{0.036 $\pm$ 0.025} & \added{0.0027 $\pm$ 0.0043} & \added{0.0401 $\pm$ 0.0008} & \added{0.150 $\pm$ 0.006} & \added{\textbf{0}}\\
    \added{BCSDM-NN}              & \added{0.329 $\pm$ 0.255} & \added{0.118 $\pm$ 0.164} & \added{0.045 $\pm$ 0.024} & \added{0.0687 $\pm$ 0.1926} & \added{0.2374 $\pm$ 0.0044} & \added{1.121 $\pm$ 0.083} & \added{646.06 $\pm$ 0.00} \\
    \bottomrule
\end{tabular}
\end{table*}
In Fig.~\ref{fig:spd2_spline}, an example is provided of a curve $\gamma_{\SPD{2}}$ fitted using the covariances of the demonstrated data in $\mathbb{R}^2$ indicated in light-blue in Fig.~\ref{fig:variable_spd_2D_rn2_vectorfield}. The effect of multiplying the velocity output of the \added{\ac{ds}}, $f_{SE(3)}$, with the transformation matrices from $\gamma_{\SPD{2}}$ is visible in the vector field of Fig.~\ref{fig:variable_spd_2D_rn2_vectorfield} on both the direction and the length of the velocity vectors. 
As we want the nominal behavior to remain unaffected, we interpolate these matrices toward identity as we approach the curve (see Fig.~\ref{fig:variable_spd_2D_rn2_vectorfield}). Inspecting the generated trajectories shows that, as the demonstrated trajectories get closer to each-other, \ie the covariance decreases, the transformed velocities induce a stronger attraction toward the curve than the original velocities. This behavior persists until the trajectory is sufficiently close to the curve, at which point the interpolation toward the identity is activated. 

%% file: sections/5_results.tex
\section{Experiments} \label{sec:experimental_results}
To verify the applicability of \ac{dsmpman}, we first provide a comparison on $S^2$ against LieFlows, PUMA\added{, and BCSDM} in Sec.~\ref{ssec: lasa_s2}, followed by an illustrative example of curve fitting in $SO(3)$ and the effect of the phase modulation layer. 
We then demonstrate \added{the proposed coupling between \ac{ds}s and curves on different manifolds in a} dressing task with a 7-axis torque-controlled Franka robot, adapting online to the configuration of a human arm, and further showcase its cross-platform applicability on a mobile manipulator dressing a mannequin arm. \added{In addition, we explore the applicability of our approach for robot-assisted shoulder rehabilitation.}

\subsection{LASA Dataset on $S^2$} \label{ssec: lasa_s2}

We compare our approach, \ac{dsmpman}, with LieFlows~\cite{urain2022learning}, PUMA~\cite{perez2024puma}\added{, and BCSDM~\cite{lee2025behavior}} on the LASA dataset~\cite{khansari2011lasa} mapped to $S^2$. 
\added{We denote the guided vector field (GVF) version of BCSDM as BCSDM-GVF, and the learned multi-task version as BCSDM-NN (DeepOVec).}
\added{%
In addition, we present a Euclidean single-chart variant of the proposed \ac{ds} as an ablation: all demonstrations are mapped once into the tangent space at the goal, the curve is fitted there with the Euclidean distance~\cite{Li25arXiv2}, rather than the manifold distance in~\eqref{eq:curve_fitting}, the \ac{ds} is constructed and integrated in the same tangent space, and only the resulting rollouts are mapped back to the manifold for evaluation.%
} The LASA dataset consists of handwritten shape trajectories. We use 26 unimodal non-intersecting shapes, each comprising 7 demonstrations. \added{All} methods are evaluated using the following metrics:
\begin{itemize}
    \item \added{\textit{RMSE}: Given a demonstration $i$ sampled with $\Delta t^i$ and initial point $x_{0}^i$, the \ac{ds} is forward simulated from $x_{0}^i$. The RMSE is the root-mean-square geodesic distance between the rollout and demonstration $i$ at the first 100 timesteps, reflecting how closely the \ac{ds} follows the demonstration in both space and time.}
    \item \added{\textit{DTWD}: The dynamic time warping distance between the forward simulations from $x_0^i$ and their corresponding demonstrations.
    It measures spatial similarity while being insensitive to differences in the speed or timing of the motion.}
    \item \textit{Path distance}: From a set of randomly sampled initial points $x_0 = \exp_{x_{\text{g}}}\left(\log_{x_{\text{g}}}(x_0^i) + \added{(2r-1)}\right)$, where $r\in [0, 1)$ and goal position $x_{\text{g}}$ lies on the manifold, the DS is forward simulated for $N=200$ timesteps and $\Delta t= 0.01 \unit{\second}$. 
    The path distance is defined as the mean geodesic distance between each point of the trajectory and the nearest discrete point in the demonstration set, and gives a measure of attraction toward the region where the demonstrations lie.
    \item \added{\textit{Goal distance}: The final geodesic distance $d_M(x(T), x_{\text{g}})$ of the randomized rollouts, with $T = N\Delta t$.}
    \item \textit{Computation time trajectory: $t_{\text{rollout}}$}. The wall-clock time required to forward-simulate a \added{single rollout from a} randomly sampled initial condition, averaged over \added{the 7 starts per shape and} all 26 shapes.
    \item \added{\textit{Query time:} The time it takes for a single evaluation of the \ac{ds}.}
    \item \textit{Train/Fitting time: $t_{\text{fit}}$}: The wall-clock time required to train the NN of LieFlows or PUMA for a single shape, or, for \ac{dsmpman}, to fit the spline and construct the DS. \added{For BCSDM-NN, it is the time of the single multi-task training that covers all 26 shapes, averaged per shape. BCSDM-GVF constructs its field directly from the reference sequence and requires no fitting.}
\end{itemize}
The curves for \ac{dsmpman} are constructed using 6 spline segments, optimized with a SLSQP solver. The initial guess is generated via a naive least-squares fit in a single tangent space, perturbed with random noise of maximum amplitude $10^{-4}$. \added{The objective~\eqref{eq:curve_fitting} is evaluated at the demonstration samples, the continuity conditions~\eqref{eq:c0_constraint_quad_bezier} and~\eqref{eq:c1_constraint_quad_bezier} enter as equality constraints, and all gradients are obtained by automatic differentiation in JAX.} For \ac{dsmpman}, the DS-parameters are set to $k_{\text{NC}} = 3.0$, $k_{\text{TC}} = 1.0$ and $k_{\text{g}} = \added{64}$. \added{
As BCSDM-GVF is defined with respect to a single data-sequence, we chose the pointwise mean of the seven demonstrations as the reference, which yielded better results than selecting any single demonstration at random. The attraction gains, $k_{\text{NC}}$ and BCSDM's $\eta$, are chosen manually,
i.e. $k_{\text{NC}} = 3$ and $\eta = 3$. For BCSDM, the speed maximum is set to twice the maximum demonstrated speed.}
For LieFlows and PUMA, training is performed for 10,000 iterations, \added{and BCSDM-NN is trained for 1000 epochs,} with all other parameters set to their default values as specified in their respective open-source implementations. 
All experiments are conducted on a standard laptop (\added{Apple M5 Pro, CPU only}).

\begin{figure}[bt]%
\centering%
\begin{subfigure}{0.48\columnwidth}%
    \centering%
    \includegraphics[width=\linewidth]{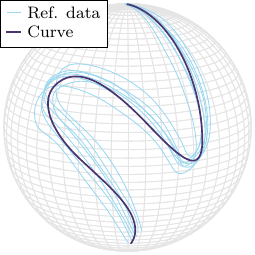}%
    \caption{}%
    \label{fig:lasa_nshape_a}%
\end{subfigure}%
\hfill%
\begin{subfigure}{0.48\columnwidth}%
    \centering%
    \includegraphics[width=\linewidth]{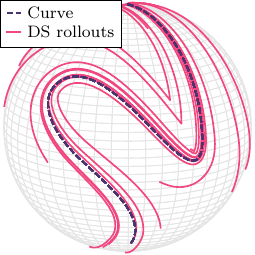}%
    \caption{}%
    \label{fig:lasa_nshape_b}%
\end{subfigure}%
\caption{%
LASA N-Shape on $S^2$. (a) Curve fitting. (b) DS rollouts.%
}%
\label{fig:lasa_nshape}%
\end{figure}
\added{
Analyzing Table~\ref{tab:lasas2}, all methods achieve low RMSE and DTWD when started from the demonstrated initial positions $x_0^i$, indicating that all methods are able to generate trajectories that are close to the demonstrated ones.
The lowest values are attained by PUMA and LieFlows, with differences to \ac{dsmpman} that are small against the spread across the 26 shapes.

Differences between the proposed \ac{dsmpman} and BCSDM-GVF arise from their reference representations: BCSDM-GVF takes a single reference trajectory, obtained here as the point-wise mean of the seven demonstrations, and reproduces its speed profile, whereas \ac{dsmpman} fits its curve to all seven demonstrations and follows the regularized speed profile of that curve.
Both methods determine the commanded speed solely from the current state, without accounting for elapsed time. Consequently, an offset introduced where the reference speed varies strongly, persists for the remainder of the rollout. Raw demonstrations accumulate more of these offsets than regularized parametrizations, leading to a higher time-indexed RMSE.
Substituting our fitted curve for BCSDM-GVF's reference sequence closes the difference,
confirming that the gap stems from the reference representation rather than from the field construction.
For more sparsely sampled demonstrations, or ones containing near-zero velocities, \ac{dsmpman} benefits from its continuous curve representation: the accuracy of the closest-point projection is independent of the sampling density, while the continuous curve representation enforces progression toward the goal.
}
\added{BCSDM-NN trades accuracy for generality, covering all 26 reference sequences with a single network, which results in improved generality at the cost of an observed performance decrease compared to BCSDM-GVF.}

\begin{figure}[tb]
\centering
\includegraphics[width=\columnwidth]{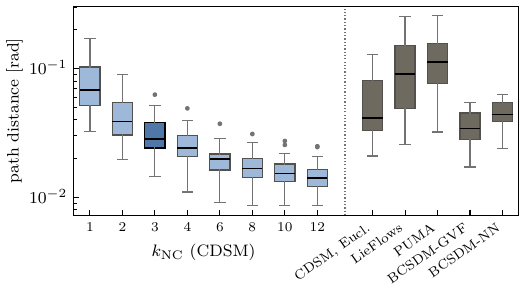}
\caption{\added{Path distance of the randomized rollouts, distributed over the 26 shapes, for \ac{dsmpman} varying the attraction gain $k_{\mathrm{NC}}$ and for the compared methods at their reported settings. The dark blue box marks $k_{\mathrm{NC}} = 3$, the value reported in Table~\ref{tab:lasas2}.}}
\label{fig:knc_path}
\end{figure}
\added{Analyzing the Path Distance in Table~\ref{tab:lasas2}, \ac{dsmpman} significantly outperforms PUMA and LieFlows, showing faster convergence toward the demonstrations.
}
This improved performance arises from \ac{dsmpman}'s ability to converge more rapidly toward the demonstration region by balancing attraction to the curve and tangential progression along the curve.
Specifically, \ac{dsmpman} explicitly modulates tangential and normal velocities, enabling faster convergence to regions with richer information and more accurate adherence to the desired movement profile.
In contrast, LieFlows occasionally fails to reach the goal within $N=200$ steps, either due to slow local progression or convergence to the antipod\added{e of the goal} $\added{-}x_{\text{g}} = (0, 0, -1)$\added{. This is reflected in the goal distances reported in Table~\ref{tab:lasas2}, where in contrast, the proposed method consistently achieves small goal distances, demonstrating reliable convergence}.
\added{The Euclidean single-chart ablation isolates the manifold-aware construction. It matches \ac{dsmpman} on the demonstration-start metrics, RMSE and DTWD, but when starting from randomized points, the rollouts approach the curve in a non-optimal, distorted way, which increases the Path Distance by more than \SI{70}{\percent} and doubles the Goal Distance.}

\added{
In Fig.~\ref{fig:knc_path}, we observe that, as expected, the Path Distance of \ac{dsmpman} decreases once we make the attraction gain $k_{\text{NC}}$ higher, since trajectories converge faster to the curve. The attraction gain is therefore an important parameter, which could be tuned by providing additional demonstrations as in~\cite{lee2025behavior}. As the convergence term is thereby proportional to the distance to the curve, it naturally ensures that the convergence term will have less effect when closer to the curve than when further from the curve.
}
All evaluated methods demonstrate real-time action querying capabilities, \added{with single-query times $t_\mathrm{query}$ of $0.026\unit{\milli\second}$ for \ac{dsmpman}, $0.072\unit{\milli\second}$ for PUMA, $0.150\unit{\milli\second}$ for BCSDM-GVF, $0.954\unit{\milli\second}$ for LieFlows, and $1.121\unit{\milli\second}$ for BCSDM-NN. \ac{dsmpman} also achieves the fastest forward simulation: for $N=200$ steps, it is approximately one order of magnitude faster than PUMA, two orders faster than LieFlows, and $24$ and $140$ times faster than BCSDM-GVF and BCSDM-NN, respectively. Its compiled rollout takes only $0.009\unit{\milli\second}$ per step, as dispatch overhead is amortized across the rollout; for the other methods, per-step rollout costs exceed their isolated query times as their rollout loops carry per-step work. For \ac{dsmpman}, projection dominates $t_\mathrm{query}$, with $\tilde{\tau}$ obtained by minimizing $d_M(x,\gamma(\tau))$ over $1000$ phase samples, giving a fixed resolution of $10^{-3}$ and accounting for \SI{20}{\micro\second} of the \SI{26}{\micro\second} query time on $S^2$. This projection cost is independent of the number and sampling density of reference trajectories.}

\added{Rollout} efficiency enables \ac{dsmpman} to provide real-time trajectory predictions that are useful for robot path planning, and also facilitates integration with high-level planners that might need to simultaneously evaluate multiple \added{\ac{ds}s} in future studies. Additionally, \ac{dsmpman} achieves rapid shape fitting in just $\added{(0.53 \pm 0.02)}\unit{\second}$ on average, making it capable of real-time DS generation. In contrast, LieFlows and PUMA require several minutes of training per shape, limiting their applicability in changing environments.
\added{BCSDM-NN is trained once for all 26 shapes, \SI{4.7}{\hour} in total, and has the highest per-step query time in the comparison, while BCSDM-GVF requires no fitting.}

\subsection{Curve Optimization on $SO(3)$} \label{ssec: curve_so3}

\begin{figure}[tb]
\centering
\includegraphics[width=\columnwidth]{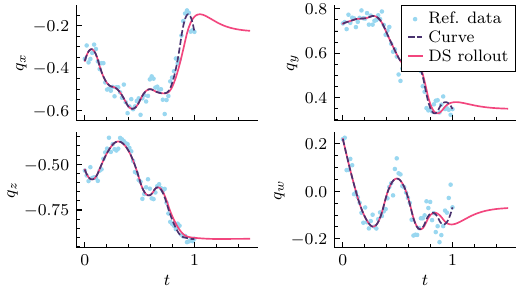}
\caption{%
Reference data, \added{fitted} curve \added{on the data}, and DS rollout on $SO(3)$, plotted over time, with $k_{\text{TC}}=1$, $k_{\text{NC}}=2$, and $k_{\text{g}}=8$.
Note the damped motion near the goal due to~\eqref{eq:twist_scaling_to_zero}.}
\label{fig:ds_time_convergence_SO3}
\end{figure}

As previously observed for LASA on $S^2$ \added{in terms of RMSE}, the proposed \added{\ac{ds}} accurately tracks both the spatial profile as well as the timing of the reference data. This is illustrated in Fig.~\ref{fig:ds_time_convergence_SO3}, which shows a forward simulation of \ac{dsmpman} on $SO(3)$ over time and highlights the effect of~\eqref{eq:twist_scaling_to_zero}, which causes a damped movement to the goal point. The parameter $k_{\text{g}}$ can be adjusted to suit different applications.
\added{The curve uses 8 segments; fitting it takes \SI{0.47}{\second} and a query of the \ac{ds} \SI{37}{\micro\second}.}

\vspace{5mm}
\subsection{Phase Modulation} \label{ssec: results_phase_modulation}
\begin{figure}
\centering
\includegraphics[width=\columnwidth]{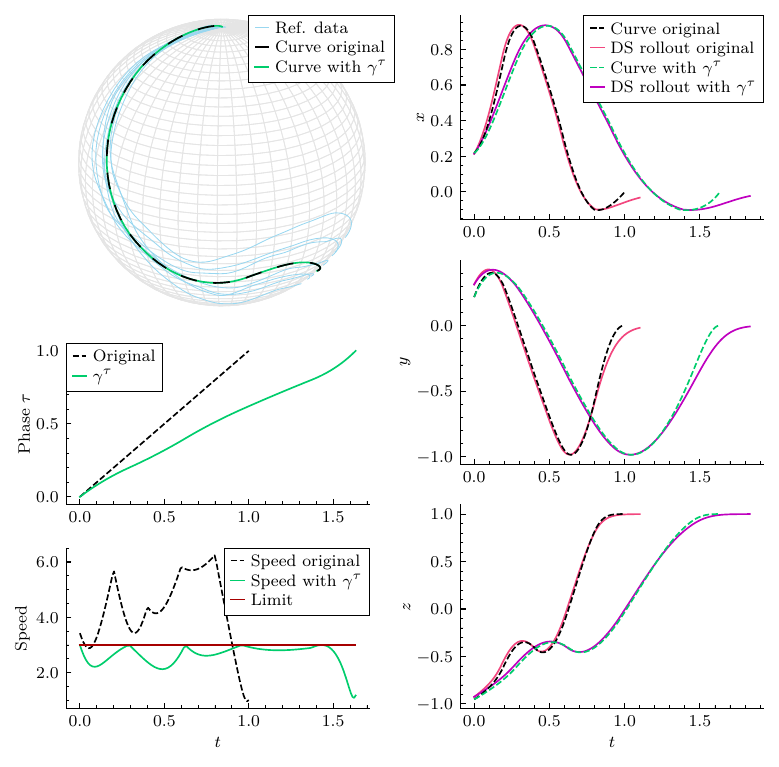}
\caption{%
The effect of an optimization of the phase modulation layer with an existing spatial curve in $S^2$ where constraints ensure that the speed, $\norm{v(t)}$, is below the limit $\bar{v} = 3.0$, and $(\gamma^{\added{\tau}})'(t) > 0$.
}
\label{fig:time_optimization}
\end{figure}

\added{The phase modulation layer enables temporal optimization while maintaining a fixed, interpretable spatial profile. Rather than being constrained by the timing exhibited during demonstration, the temporal evolution can be optimized independently to satisfy task-specific objectives and constraints.}
In Fig.~\ref{fig:time_optimization}, a phase modulation profile is shown which optimizes the time to reach the goal while adhering to speed constraints, $\norm{v(t)} \leq \bar{v}$, with the velocity $v(t)$ defined in the local tangent space $T_{x(t)}S^2$ and limit $\bar{v} = 3.0$. The phase modulation curve, $\gamma^{\added{\tau}}$, consists of \added{5} segments. 
To prevent backward motion along the spatial profile, we enforce a constraint ensuring that the derivative of the phase curve, $(\gamma^{\added{\tau}})'(t) = \frac{\delta \added{\tau}}{\delta t}$, remains strictly positive.
As observed in Fig.~\ref{fig:time_optimization}, the spatial profile of the curve on $S^2$ is preserved, while the temporal profile is modified to ensure that the speed remains below the prescribed maximum, in contrast to the original speed profile. Comparing the phase-modulated curve $\gamma^{\added{\tau}}$ with the original curve shows that not only the time to reach the goal, but also the velocity profile along the curve and of the DS rollouts is altered. 
The introduction of the phase modulation layer preserves the system's numerical stability while effectively modifying the trajectory characteristics. Computation times for fitting the curve $\gamma^{\added{\tau}}$ are \added{$(58 \pm 3)\unit{\milli\second}$} averaged over all 26 $S^2$-shapes using \added{5} segments, and \added{$(26.9 \pm 0.3)\unit{\micro\second}$} for querying an action \added{of the modulated \ac{ds}, essentially the same as for the unmodulated \ac{ds}}. 
Future work could explore extending the phase modulation to enable collision avoidance in dynamic environments, as well as learning phase modulation profiles directly from real-world data.

\subsection{Human Arm Dressing} \label{ssec: results_dressing}
\begin{figure}[tb]
\centering
\includegraphics[width=\columnwidth]{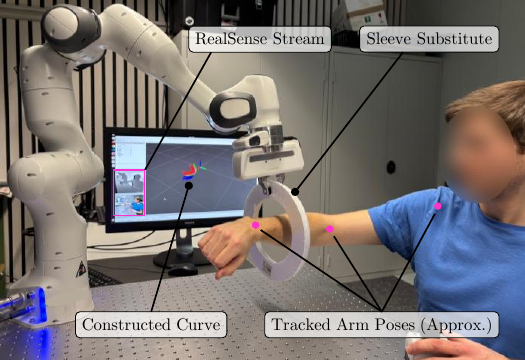}
\caption{%
Setup of the arm dressing experiment. The RealSense camera is positioned near the viewpoint of this photo.
}
\label{fig:experiment_setup}
\end{figure}

To evaluate the real-world applicability of the proposed DS and variable damping framework, we deploy our method on a robotic dressing task (see Fig.~\ref{fig:experiment_setup}). From previous studies addressing such task, \eg~\cite{wangOnePolicyDress2023, franzese_generalizable_2025}, we conclude that this scenario demands fast and continuous online adaptation, as the system must simultaneously ensure human-safe compliance, track the human arm online, and achieve the precision necessary for garment manipulation.

\subsubsection{Experimental Setup and Sensing}

We run the experiments on a Franka robot, utilizing Cartesian impedance control.
The control loops for high- and low-level commands run at \SI{1}{\kHz}. \added{Within this loop, a query of $f_{SE(3)}$ takes \SI{65}{\micro\second} and the evaluation of the damping curve $\gamma_{\SPD{6}}$ at the corresponding phase \SI{45}{\micro\second}; the spline itself is reconstructed in closed form from the three tracked poses once per camera frame.} For arm tracking, we use an Intel RealSense D415 camera and perform initial processing with MediaPipe~\cite{lugaresiMediaPipeFrameworkBuilding2019}. We extract the 3D coordinates of the user's wrist, elbow, and shoulder to construct a composite quadratic \Bezier spline with two segments online. This spline defines the trajectory of end-effector poses in $SE(3)$ along which the robot follows the arm.

\subsubsection{Controller Formulation}

The following variable impedance control law governs the interaction:
\begin{subequations} \label{eq:impedance_control_law}
\begin{align}
    \added{\mu}(t) &= J^\transpose \, D_c \, \Bigl( D(x) \ \dot{x}_{d} - \dot{x} \Bigr) \label{eq:impedance_control_law_a}\\
    &= J^\transpose \, D_c \, \Bigl( \gamma_{\SPD{n}}\!\bigl(\tilde{\added{\tau}}(x\bigr)\bigr) \ f_{SE(3)}\!(x) - \dot{x} \Bigr), \label{eq:impedance_control_law_b}
\end{align}
\end{subequations}
where $\added{\mu}(t)$ is the commanded torque, $J$ is the geometric Jacobian, $D_c \in \SPD{6}$ represents the fixed damping gain of the Cartesian controller, $\dot{x}_d$ and $\dot{x}(t)$ are the desired and current twist, respectively, and $f_{SE(3)}$ is the nominal DS output. The variable damping matrix $D(x(t))$ is updated at \SI{1}{\kHz} alongside the DS, and applied to the desired twist.
For fixed damping, we simply omit $D(x(t))$.
This control law is augmented with gravity compensation and null-space control to keep the robot in singularity-free regions of the workspace.
The parameters of the $f_{SE(3)}$ are chosen to be $k_{\text{TC}} = 1$, $k_{\text{NC}} = 2$, $k_{\text{g}} = \added{16}$, $L_c=0.01$ for closest point projection, and $L_c=0.1$ for any other operations involving distance computation.

\begin{figure}[tb]
\centering
\includegraphics[width=\columnwidth]{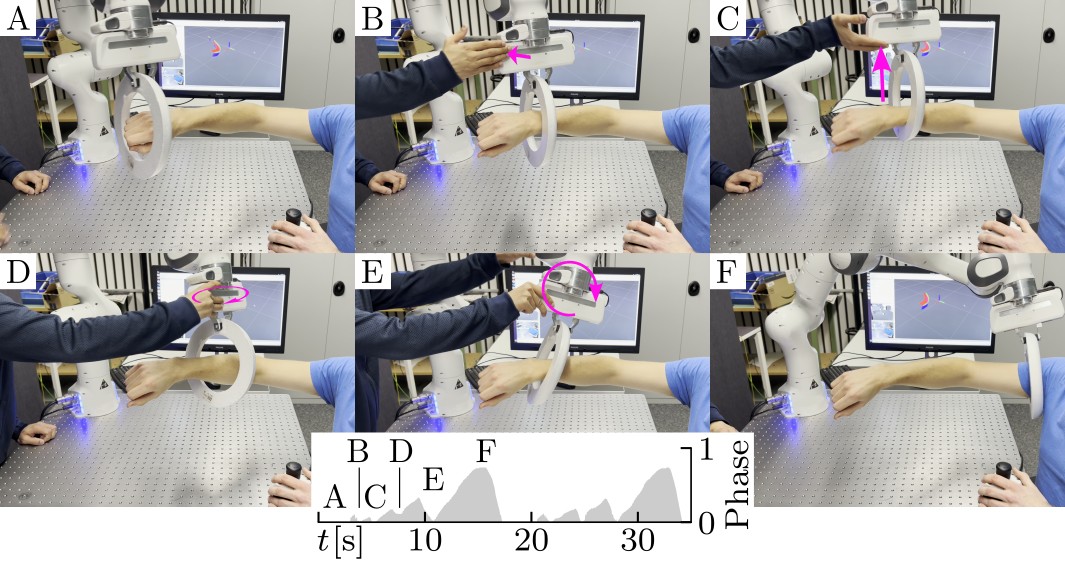}
\caption{%
Various snapshots of the scene during the dressing task.
Frames A and F show the beginning and end of the first of two repeated runs, respectively.
The other frames show different kinds of perturbations: B push laterally, C push up, D twist counter clockwise, E tilt forward.
}
\label{fig:experiment_frames}
\end{figure}
\begin{figure}[tb]
\centering
\includegraphics[width=\columnwidth]{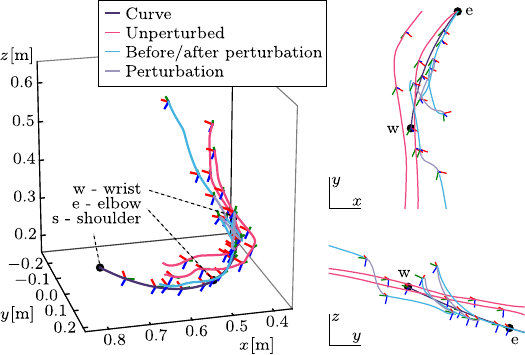}
\caption{%
A snapshot of the curve for a tracked steady arm (dark violet), \added{extracted from tracking the wrist, elbow and shoulder,} with two trajectories of the robot when not pushed (pink). We also show some perturbations (gray) and the behavior of the robot \added{before and} after the perturbations (light blue).
The world frame is aligned with the robot base frame.
Due to the human arm not being perfectly still, the curve moved a little throughout the experiment, and hence the illustrated curve is only a snapshot.
}
\label{fig:experiment_SE3}
\end{figure}
\subsubsection{Experimental Cases}

\begin{figure}[tb]
\centering
\includegraphics[width=\columnwidth]{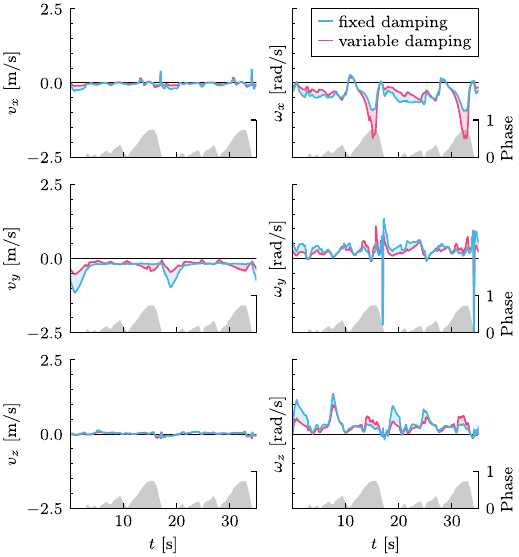}
\caption{%
Robot moving along a static arm with perturbations to position and orientation, repeated twice.
The experiment is performed with the damping varying linearly with the phase along the nominal curve from 0.5 to 2. The resulting twist outputs of the DS are plotted in pink, the phase in gray and the twists when using a fixed damping are in blue.
\added{Note that the jumps in this figure and Fig.~\ref{fig:experiment_torques_over_time}, are caused by curve estimation jitters due to camera occlusions, and the decrease in phase by the perturbations of the second human.}
}
\label{fig:experiment_twists_over_time}
\end{figure}

We consider five scenarios: (i) dressing a static arm, (ii) perturbations before the garment (represented by the white ring) is around the arm, (iii) perturbations during movement along a static arm, (iv) user-induced perturbations during execution, in which the robot is physically displaced while the user simultaneously moves their arm to a different configuration, and (v) dressing a moving arm, where the user actively adjusts their wrist, elbow, and shoulder poses in real time.
Here, for visual interpretability, variable damping matrices are constructed as a curve of diagonal matrices in $\SPD{6}$, though our framework supports full $\SPD{6}$ matrices. During the experiment, the gains in $D(x(t))$ vary smoothly from
0.5 to 2 based on the progress along the spline. This results in an intentional increase in velocity gains as the end-effector approaches the shoulder.
As mentioned in Sec.~\ref{ssec:pose_ds_damping}, the variable damping matrices obtained from $\gamma_{\SPD{n}}$ were applied fully when the distance from the curve was greater than a threshold of 0.05 for the experiment in Fig.~\ref{fig:experiment_SE3} and 0.01 for the experiment in Figs.~\ref{fig:experiment_frames}, \ref{fig:experiment_twists_over_time}, \ref{fig:experiment_torques_over_time}, and interpolated toward identity when closer to the curve.

For the plots presented in this paper, we focus on the scenario of a stationary human arm, while introducing perturbations by pushing or pulling the robot during the dressing motion. This analysis includes both fixed and variable damping conditions. Fig.~\ref{fig:experiment_frames} shows snapshots of an experiment using a DS in $SE(3)$ with variable damping, captured at key time points. Additionally, the supplementary video demonstrates all five scenarios, where a static as well as a moving arm is dressed.
\subsubsection{Human Arm Dressing with a Mobile Manipulator}

To verify the applicability of our approach across different robotic platforms, and to observe its performance in the presence of internal perturbations caused by the physical limitations of the robotic system, we demonstrate it on a mobile manipulator comprising a Kinova Gen3 Lite mounted on a Clearpath Dingo base (see Fig.~\ref{fig:dinova_experiment}). Given that the mobile manipulator only accepts joint velocity commands, the poses generated by $f_{SE(3)}$ are directly fed into the low-level geometric controller~\cite{spahn2023dynamic}. Illustrative experiments were conducted using several arm poses of a mannequin arm and starting poses of the robot. Joint velocity commands and high-level commands are sent at \SI{40}{\Hz} and \SI{20}{\Hz} respectively, and the arm coordinates and robot base pose are tracked online with a VICON motion capture system. 

\subsection{\added{Human-Arm Dressing: }Results and Discussion}
\begin{figure}[tb]
\centering
\includegraphics[width=\columnwidth]{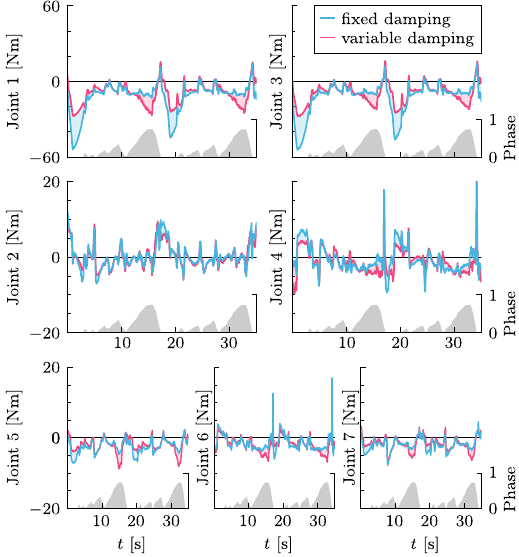}
\caption{%
Resulting torque values of the twists applied in Fig.~\ref{fig:experiment_twists_over_time} over time, with the phase in gray. The torque commands from~\eqref{eq:impedance_control_law} with $D_c=\mathrm{diag}(90,90,90,4,4,4)$ are shown in pink, while the torque commands with a fixed damping are in blue.
}
\label{fig:experiment_torques_over_time}
\end{figure}
In Fig.~\ref{fig:experiment_SE3}, the robot approaches the static arm
from some distance away and continues to follow the curve, which is continuously updated based on the tracked arm pose. Imperfect alignment with the curve is caused by various factors: (a) the arm is not perfectly still, (b) contact between the foam ring and the arm due to inaccuracies in the spline offsets used for its construction, and (c) camera noise on the arm pose tracking.
Note that (a) is a significant factor in Fig.~\ref{fig:experiment_SE3}, which directly explains why the perturbed trajectories appear to converge to the curve quicker than the unperturbed ones. The curve shown in Fig.~\ref{fig:experiment_SE3} is a snapshot of the curve when perturbations were applied, and the curve varied a little throughout the experiment due to human arm movements.
We further note that for the experiment the robot is unable to reach the very end of the curve because it is impeded by the armpit, as observed in Fig.~\ref{fig:dressing}.
When external perturbations are applied, observed as gray lines in Fig.~\ref{fig:experiment_SE3}, the system complies and continues to follow the DS afterward, converging to the curve, as indicated in light blue. 

Next, we analyze the effect of variable damping on the twists, joint torques, and human-observed stiffness of the movement.
In Figs.~\ref{fig:experiment_twists_over_time} and~\ref{fig:experiment_torques_over_time}, the twist and joint torque values are shown with fixed and variable damping, where the perturbation sequence as in frames A-F in Fig.~\ref{fig:experiment_frames} is repeated twice.
Joint torques under variable damping are applied to the robot, while those under fixed damping are derived for comparison, because identical human disturbances cannot be replicated.
When variable damping is enabled, the twist outputs from $f_{SE(3)}$ are modified based on the current phase along the nominal curve. In the particular example of Fig.~\ref{fig:experiment_twists_over_time}, a linear scheduling of the damping gain from 0.5 to 2 across the phase leads to lower twists early on and increased twist magnitudes at higher phases. 
The increased damping strength from wrist to shoulder renders the robot softer and less aggressive during the approach phase and during early stages of the nominal curve, and stiffer and more reactive
closer to the shoulder, as observed from the command torque values in Fig.~\ref{fig:experiment_torques_over_time} and perceived by the users.

During experiments, occasional jumps in the tracked curve were observed due to the inherent limitations of the vision system and occlusions.
However, due to the fast recomputation of our system and the compliant nature of our controller,
these discontinuities did not result in excessively jerky motions or safety hazards for the human subject. Whenever tracking confidence reported by MediaPipe was below 0.9, the last good arm pose estimate was used.

\begin{figure} [tb]
    \centering
    \includegraphics[width=\linewidth]{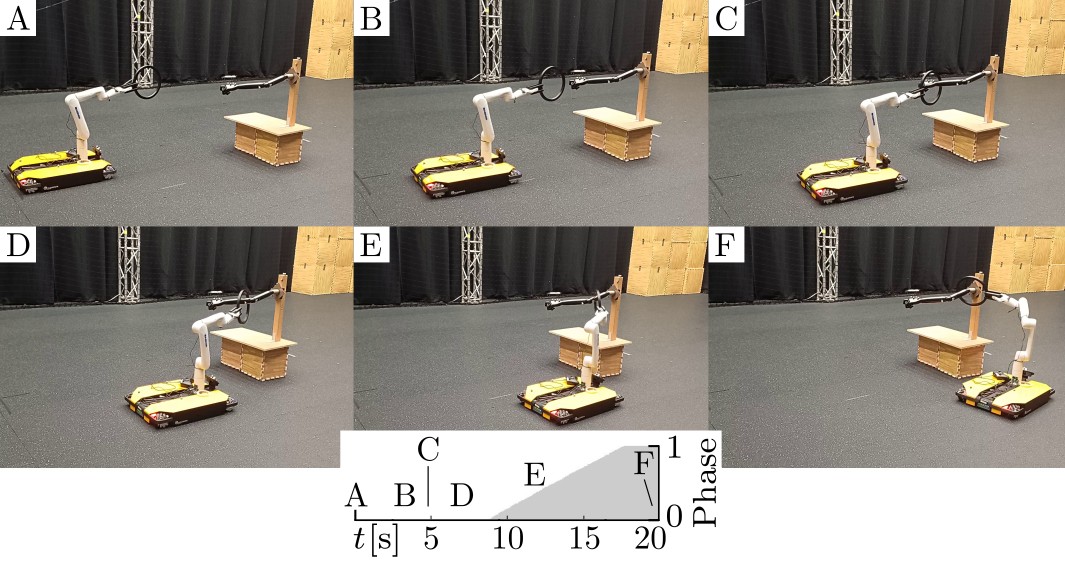}
    \caption{The dressing task executed by a mobile manipulator consisting of a Kinova Gen3Lite mounted on a Clearpath Dingo base. Frames A and F indicate the start and end-pose, respectively.}
    \label{fig:dinova_experiment}
\end{figure}

On the mobile manipulator, the proposed \ac{dsmpman} executes the dressing task of a mannequin arm from different initial positions of the robot's base and configurations of the mannequin arm, as observed in the video and Fig.~\ref{fig:dinova_experiment}. 
When reaching the elbow joint of the mannequin, the mobile manipulator must reconfigure both its mobile base and arm before continuing the motion of the end-effector. This demonstrates the system's ability to handle not only external disturbances (Fig.~\ref{fig:experiment_frames}), but also internal perturbations caused by the physical constraints of the robotic platform, emphasizing the advantage of \ac{dsmpman} over a time-based motion primitives.
Due to wheel slip with the omnidirectional wheels, as well as additional damping induced by the low-level controller, the tracking accuracy was diminished with respect to the experiments on the Franka robot. Incorporating visual feedback on the current ring or garment placement with respect to the human arm could make the dressing task more realistic for users~\cite{wangOnePolicyDress2023}.
Nevertheless, the experiment highlights our framework's cross-platform capabilities and its potential use in real-world at-home scenarios.

\subsection{\added{Robot-assisted shoulder rehabilitation}} \label{ssec: robot_therapy}
\begin{figure}
    \centering
    \includegraphics[width=\linewidth]{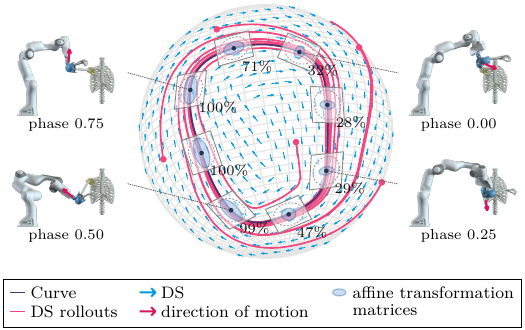}
    \caption{
    \added{
    The coordinates on $S^2$ represent the plane of elevation and shoulder elevation, respectively. \ac{dsmpman} defines a \ac{ds} on $S^2$ that guides the human toward and along the desired exercise path, as indicated by the curve. This is coupled via phase to a curve of affine transformation matrices, which model the assistance-as-needed during the exercise. The percentage reflects the level of assistance, with 100\% indicating maximum assistance and 0\% indicating no assistance. The skeleton is generated using the biomechanical model described in~\cite{seth2018opensim, seth2019muscle}.
    }
    }
    \label{fig:cyclic_shoulder}
\end{figure}

\begin{figure}
    \centering
    \includegraphics[width=0.8\linewidth,trim={0 0 0 1.14cm}, clip]{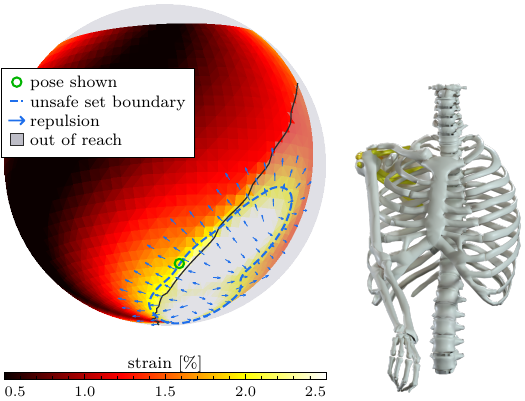}
    \caption{\added{
    The repulsive field from a high-strain region of the shoulder tendons, where the curve models the boundary of the high-strain region.
    The passive strain map and skeleton pose are generated using the biomechanical model~\cite{seth2018opensim, seth2019muscle}. The skeleton pose corresponds to the pose on $S^2$ in green. 
    }}
    \label{fig:strain_map}
\end{figure}

\added{Another relevant application of \ac{dsmpman} is robot-assisted shoulder rehabilitation~\cite{belli2025shared}, as illustrated in Fig.~\ref{fig:cyclic_shoulder} and~\ref{fig:strain_map}.
Shoulder and rotator cuff injuries affect up to 22\% of the general population~\cite{minagawa2013prevalence}, and rehabilitation typically requires a physiotherapist to guide the patient's movements or the use of assistive devices, such as elastic bands~\cite{thomson2016rehabilitation, schedler2020effects}.
Robot-assisted rehabilitation can support these exercises by providing guiding motions while adapting the stiffness and damping of the impedance controller according to the desired rehabilitation strategy. To ensure safe motions, regions of high tendon strain should be avoided, which can be identified using a biomechanical model~\cite{seth2018opensim, seth2019muscle}. Moreover, rehabilitation exercises are typically performed repetitively and need to be adaptable to individual patients and to changes in their capabilities throughout the rehabilitation process. In this section, we demonstrate how our \ac{dsmpman} can generate cyclic exercise motions, adapt assistance-as-needed throughout the motion, and guide the human away from high-strain regions.}

\added{In Fig.~\ref{fig:cyclic_shoulder}, \ac{dsmpman} defines a cyclic shoulder exercise on $S^2$, where the two coordinates represent the plane of elevation and shoulder elevation, respectively. The cyclic curve represents the desired exercise, and the \ac{ds} guides the human toward and along the desired exercise. The \ac{ds} on $S^2$ is coupled with a varying affine transformation matrix along the exercise to modulate the assistance-as-needed.
In a robot-assisted deployment, the same control law as~\eqref{eq:impedance_control_law_a} applies with $\dot{x}_d=f_{S^2}(x)$ and $D(x)=\gamma_{\SPD{2}}$, with $J$ the robot's Jacobian mapping to the task space in which the exercise is executed.
As shown in Fig.~\ref{fig:cyclic_shoulder}, the resulting assistance is greater during the upward motion (phase $\in[0.5,0.75]$) than during the downward motion (phase $\in[0.0,0.25]$). Here, 0\% assistance indicates that the human receives no assistance in the exercise direction, whereas 100\% corresponds to maximum assistance in that direction. The varying affine transformation matrices and exercise motion are currently defined manually, but could instead be learned from physiotherapist demonstrations~\cite{luciani2023trajectory}.}

\added{
In addition, we can leverage \ac{dsmpman} by defining an enclosed high-strain region of the shoulder tendons as a cyclic curve, as shown in Fig.~\ref{fig:strain_map}. We slightly modify the \ac{ds} of \ac{dsmpman} to define a repulsive instead of contracting \ac{ds},
\begin{align}
    \dot{x}_{\text{RP}} &= f_{\text{RP}} = k_{\text{NC}} k_{\text{RP}} \frac{\nabla  \Psi_{\text{NC}}(x)}{||\nabla  \Psi_{\text{NC}}(x)||},
\end{align}
where $k_{\text{RP}} = -1$, when inside the unsafe region, and $k_{\text{RP}} = \max\left(0, \min\left(1, 1-\frac{d_{M}}{d_{\text{max}}}\right)\right)$, otherwise. We normalize $\nabla  \Psi_{\text{NC}}(x)$ to ensure that the magnitude of the vector does not vanish when close to the boundary of the unsafe strain-region $\mathcal{S}_{\text{unsafe}}$, and on the boundary establish a numerical gradient pointing toward the safe region. The scaling-factor $k_{\text{RP}}$ ensures that within the unsafe region, the repulsive term directs the \ac{ds} toward the safe region, while in the safe region, its effect is highest close to the boundary and decreases toward zero at a maximum distance $d_{\text{max}}$ to the boundary. Note that, if multiple distinct unsafe regions are present, this requires pre-processing to establish the boundary data-points of each enclosed unsafe region.
We incorporate $\dot{x}_{\text{RP}}$ as a damping on the current velocity into the impedance controller, }
\added{\begin{equation} \label{eq:impedance_control_law_strain}
    \added{\mu}(t) = J^\transpose \, D_c \, \Bigl( D(x(t)) \ \dot{x}_{d} - D_{\text{strain}}\dot{x}(t) \Bigr),
\end{equation}}%
\added{with $D_{\text{strain}} = I + \text{max}(0, - \langle \dot{x}, \dot{x}_{\text{RP}} \rangle) n_{\text{RP}} n_{\text{RP}}^{\top}$, and $n_{\text{RP}} = \frac{\dot{x}_{\text{RP}}}{||\dot{x}_{\text{RP}}||+ \epsilon}$}.
\added{The damping scaling $D_{\text{strain}}$ is zero when moving away from the unsafe set, and increases magnitude closer to the unsafe set and when the current velocity toward the unsafe set increases, in contrast to the static damping in the damped zone defined in~\cite{belli2025shared}. $D_{\text{strain}}$ therefore provides the human with feedback on avoiding the high-strain region.
Real-world experiments involving physiotherapists could provide a valuable next step toward validating the applicability of \ac{dsmpman} in rehabilitation.
}

%% file: sections/7_conclusion.tex
\section{Limitations and Conclusion}
We proposed \ac{dsmpman}, a framework for \added{real-time generation and coupling of motion profiles on Riemannian manifolds and Lie groups}.
\added{Building a smooth phase-parametrized curve on the manifold, \ac{dsmpman} defines a \ac{ds} that combines an attracting normal vector toward a curve and a propagating tangential vector in the direction of the curve. We leverage the associated phase in three ways: first, to establish stability within the local analysis domain and to analyze crossings of ridges of the distance potential, where the closest-point projection is ambiguous; second, to couple \ac{ds}s and curves that live on different manifolds; and third, to independently modulate the timing of the motion.}
\added{Quantitative experiments show that \ac{dsmpman} outperforms LieFlows and PUMA on an $S^2$ dataset in terms of path and goal distance, as well as generation and query time, while achieving comparable path distance and higher reproduction accuracy than BCSDM-GVF. This similarity is expected since BCSDM-GVF follows a closely related \ac{ds}, only attracting to the demonstrated trajectory itself. In contrast, \ac{dsmpman} leverages the continuous-time curve to define phase-dependent coupling and a phase-modulation layer for time-optimization under constraints.}
\added{We demonstrate the applicability of coupling a spatial \ac{ds} with varying damping matrices of an impedance controller for a dressing task on both a real-world manipulator and a mobile manipulator, as well as the suitability of \ac{dsmpman} for robot-assisted shoulder rehabilitation.}

Our current experiments are limited to uni-modal demonstrations and movements, but the proposed approach naturally extends to multimodal scenarios.
One possible extension involves defining a single DS from multiple curves and simply projecting toward the closest curve for any given query point.
Alternatively, multiple DSs, each associated with its own curve, could be generated, with the most relevant DS selected dynamically based on the observed scenario.
Additionally, we introduced a coupling between a pose-based \added{\ac{ds}} and varying $\SPD{n}$ matrices representing the velocity gains. In future work, these matrices and the parameters of the \added{\ac{ds}} could be learned from human data~\cite{ajoudani2015reduced}, enabling adaptable and human-like movements.
When adapting the damping matrix online, we assume gradual changes to preserve the stability of the closed-loop robotic system. Under this assumption, a passivity analysis could be performed to verify stability during damping adaptation.
To further enhance system robustness, we suggest introducing an energy-budget constraint for the Cartesian impedance controller~\cite{kronanderPassiveInteractionControl2016, lachner2021energy}. This constraint would cap the maximum potential energy of the system, ensuring that the proportional term of the controller only steers toward poses that comply with the energy limit, thereby safeguarding against instability.

%% file: sections/8_appendix.tex
\section{\added{Stability Analysis}} \label{sec: appendix_proof}
\added{This Appendix establishes the convergence properties of the proposed dynamical system. In App.~\ref{ssec: part1_local_to_curve}, we show that the system converges to the curve using a distance-based Lyapunov candidate on the local domain $\mathcal{D} = M \setminus \mathcal{C}_{\text{sing}}$. In App.~\ref{ssec: part2_local_conv_goal}, we extend this analysis to establish local convergence to the goal point $\gamma(1)$ by incorporating the tangential potential into the Lyapunov candidate. Finally, App.~\ref{ssec: part3_convergence_local_max} considers trajectories in the presence of the singular set $\mathcal{C}_{\text{sing}}$, addressing the possible discontinuities arising from multiple closest-point projections and showing that the energy decreases in the presence of crossings of the singular set.}

\subsection{\added{Local convergence to the curve}} \label{ssec: part1_local_to_curve}
\added{To analyze \textbf{local convergence to the curve} on the domain $\mathcal{D}$, we consider the Lyapunov function defined as the squared minimum geodesic distance to the trajectory. Equivalently, this can be expressed as the energy of the minimizing geodesic $\eta_x$,
\begin{equation} \label{eq: lyapunov_conv_curve}
\small{V(x) =
\Psi_{\text{NC}}(x)
=
\frac{1}{2} d_{M}^2(\gamma, x) =
\frac{1}{2}
\int_0^1
\left\langle
\eta_x'(s),
\eta_x'(s)
\right\rangle ds,}
\end{equation}
where $\eta_x\colon [0,1]\rightarrow \mathcal{D}$ is the minimizing geodesic satisfying
$\eta_x(0)=x$ and
$\eta_x(1)=\gamma(\tilde{\tau}(x))$, with $\tilde{\tau}$ as in~\eqref{eq:projection}.
The tangent vector along the geodesic is given by
$\eta_x'(s) = \frac{d\eta_x}{ds}(s)$. Note that the variable $s$ indicates the ``phase'' along the geodesic $\eta(s)$, and $\tau$ the phase along the curve $\gamma(\tau)$.}

\added{
The objective is to show that the \ac{ds} trajectories converge to the curve $\gamma(\tau)$. This requires the Lyapunov function to decrease along the \ac{ds} trajectories. Differentiating $V(x)$ in~\eqref{eq: lyapunov_conv_curve} with respect to time yields
\begin{equation}
\frac{dV(x)}{dt}
=
d_{M}(\gamma, x)\dot{d}_M(\gamma, x)
=
\int_0^1
\left\langle
\dot{\eta}_x(s),
\eta_x'(s)
\right\rangle
\,ds,
\end{equation}
where $\dot{\eta}_x(s)=\frac{d}{dt}\eta_x(s)$ denotes the time derivative of the minimizing geodesic.
Applying integration by parts gives
\begin{equation}\label{eq:proof_int_parts}
\begin{aligned}
\frac{dV(x)}{dt}
=
& \left\langle
\eta_x'(1),
\dot{\eta}_x(1)
\right\rangle
-
\left\langle
\eta_x'(0),
\dot{\eta}_x(0)
\right\rangle \\
&\quad -
\int_0^1
\left\langle
\dot{\eta}_x(s),
\nabla_{\eta_{x}'} \eta_x'(s)
\right\rangle
\,ds.
\end{aligned}
\end{equation}
Since $\eta_x$ is a geodesic, its covariant acceleration vanishes,
$\nabla_{\eta_{x}'} \eta_x'(s) = 0,$
and therefore the final integral vanishes.}

\added{First, consider the \textbf{first inner product} in~\eqref{eq:proof_int_parts}, $\left\langle
\eta_x'(1),
\dot{\eta}_x(1)
\right\rangle$, which corresponds to analyzing the endpoint of the geodesic.
Since the endpoint of the geodesic coincides with the point on the curve, the time derivative of the endpoint is the velocity of the curve. We introduce $c = \frac{d\tilde{\tau}}{dt} \in \mathbb{R}$, to express the time derivative of the endpoint in terms of the derivative of the curve, $\gamma'(\tilde{\tau})$. Hence, 
\begin{equation} \label{eq: gamma_dt}
    \dot{\eta}_x(1) = \frac{d}{dt} \gamma(\tilde{\tau}(x)) = c \gamma'(\tilde{\tau}(x)).
\end{equation}
Three cases arise depending on whether the geodesic endpoint lies in the curve interior, $\tau \in (0,1)$, at an endpoint, $\gamma(1)$, or the beginning of the curve, $\gamma(0)$:}

\added{\textbf{Case 1:} When the closest point on the curve is in the interval $\tau \in (0, 1)$, the curve derivative $\gamma'(\tilde{\tau}(x))$ and minimizing geodesic tangent vector $\eta_x'(1)$ are strictly orthogonal by the First Variation of Arc Length (Gauss's Lemma)~\cite{carmoRiemannianGeometry1992}, hence $\left\langle
\eta_x'(1),
\dot{\eta}_x(1)
\right\rangle = 0$.}

\added{\textbf{Case 2:} When the closest point on the curve is at the goal $\tilde{\tau} = 1$, we defined the velocity along the curve to be zero in~\eqref{eq:twist_scaling_to_zero}, $\frac{d}{dt}\gamma(1)= 0 $, rendering the inner product to zero.}

\added{\textbf{Case 3:} The most challenging case occurs when the closest point is at the curve's beginning, i.e., $\tilde{\tau}=0$, where orthogonality is not always guaranteed, as illustrated in Fig.~\ref{fig:lasa_ridges}. For points projecting towards $\gamma(0)$, it holds that,
\begin{equation}
\left\langle
\eta_x'(1), \gamma'(0)
\right\rangle \geq 0,
\end{equation}
since $\tilde{\tau}=0$ corresponds to a local one-sided minimum of the distance $d_M$ to the curve on $[0, 1]$.
If the tangent vector of the minimizing geodesic and the curve derivative are orthogonal, the inner product vanishes by Gauss' lemma. When orthogonality is not established, a set of points remains for which $
\left\langle
\eta_x'(1), \gamma'(0)
\right\rangle > 0$.}

\added{For points with $
\left\langle
\eta_x'(1), \gamma'(0)
\right\rangle > 0$, we leverage Lemmas 1 and 2 in~\cite{lee2025behavior}, to conclude that
$
\dot{\eta}_x(1) = 0,
$
which implies
$
\left\langle
\eta_x'(1), \dot{\eta}_x(1)
\right\rangle = 0.
$
By Lemma 1 in~\cite{lee2025behavior}, $\tilde{\tau}$ is continuous in $x$. Consequently, both $\gamma(\tilde{\tau})$ and $\eta_x'(1)$ are continuous in $x$. Moreover, all points for which $
\left\langle
\eta_x'(1), \gamma'(0)
\right\rangle > 0$,
project onto the initial point of the curve, corresponding to $\tilde{\tau}=0$. In the local neighborhood where $\tilde{\tau}=0$, Lemma 2 in~\cite{lee2025behavior} gives $\frac{d \tilde{\tau}}{dt}=0$. Therefore, \eqref{eq: gamma_dt}~reduces to $
\frac{d}{dt}\gamma(0) = 0$.
Hence, $
\left\langle
\eta_x'(1), \dot{\eta}_x(1)
\right\rangle = 0$.
For a more extensive analysis of this third case, we refer to App. A in~\cite{lee2025behavior}.}
\added{
The three cases above establish that the first inner product is zero, $\left\langle
\eta_x'(1),
\dot{\eta}_x(1)
\right\rangle = 0$.}

\added{Next, we analyze the \textbf{second inner product} in~\eqref{eq:proof_int_parts}, which is evaluated at the beginning of the minimizing geodesic. At this point, the time derivative of the geodesic satisfies
$\dot{\eta}_x(0)=\frac{d}{dt}x(t)$,
since $\eta_x(0)$ corresponds to the current state $x(t)$. Therefore,~\eqref{eq:proof_int_parts} can be rewritten as
\begin{equation}
\frac{dV(x)}{dt}
=
-
\left\langle
\eta_x'(0),
\dot{x}
\right\rangle .
\end{equation}}
\added{Inserting the \ac{ds}, $\dot{x} = f(x)$ from~\eqref{eq:ds},
\begin{subequations} \label{eq: conv_to_curve_dvdt_1}
\begin{align}
\frac{dV(x)}{dt}
=
&-
\left\langle
\eta_x'(0),
f_{\text{prop}} + f_{\text{conv}}
\right\rangle\\
=
&- \langle
\eta_x'(0),
k_{\text{TC}} \PT{\proj(x)}{x}\left(\gamma'\left(\tilde{\tau}\right)\right)\rangle 
\notag \\
&\quad - 
\left\langle
\eta_x'(0), - k_{\text{NC}} \nabla \Psi_{\text{NC}}(x)\vphantom{\PT{\proj(x)}{x}}
\right\rangle.
\end{align}
\end{subequations}}
\added{Since $- \nabla \Psi_{\text{NC}}(x)\vphantom{\PT{\proj(x)}{x}}$  is defined as the tangent velocity vector of the geodesic connecting $x$ to $\pi(x)$, i.e. $- \nabla \Psi_{\text{NC}}(x)\vphantom{\PT{\proj(x)}{x}} = \eta'(0)$, and $\norm{\eta'(0)} = d_M$, we obtain
\begin{align}
\frac{dV(x)}{dt} &=
- k_{\text{TC}}
\langle
\eta_x'(0), \PT{\proj(x)}{x}(\gamma'(\tilde{\tau}))\rangle -
k_{\text{NC}} d_M^2 .
\end{align}
We leverage the parallel transport from $x$ to $\pi(x)$, $\PT{x}{\proj(x)}$, to establish
\begin{equation} \label{eq: dvdt_proof_54}
\frac{dV(x)}{dt} =
-
k_{\text{TC}}
\left\langle
\eta_x'(1), \gamma'(\tilde{\tau})\right\rangle -
k_{\text{NC}} d_{M}^2,
\end{equation}
as the parallel transport preserves the inner product explicitly.}

\added{The first part, $\left\langle
\eta_x'(1), \gamma'(\tilde{\tau})\right\rangle$, can again be analyzed via the three cases, $\tilde{\tau} \in (0, 1)$, $\tilde{\tau} = 1$ and $\tilde{\tau} = 0$.
In Case 1, due to orthogonality of $\eta'_x(1)$ and $\gamma'(\tilde{\tau}(x))$, the inner product is zero when $\tilde{\tau} \in (0, 1)$.
In the second case, $\tilde{\tau} = 1$, we establish that $\gamma'(1)=0$, rendering the first inner product to zero. 
For the third case, the inner product between $\eta_x'(1)$ and the curve derivative $\gamma'(\tilde{\tau}(x))$ are nonnegative,
\begin{equation}
\left\langle
\eta_x'(1), \gamma'(\tilde{\tau})
\right\rangle \geq 0, \ \text{if} \ \tilde{\tau} = 0.
\end{equation}
The derivative of the Lyapunov function therefore becomes
\begin{subequations} \label{eq: dvdt_curve_attract_conclude}
\begin{align}
\frac{dV(x)}{dt} &= - 
k_{\text{NC}} d_M^2, \ \ \text{if } \ \tilde{\tau} \in (0, 1],\\
\frac{dV(x)}{dt} &\leq -  k_{\text{NC}} d_M^2, \ \ \text{if } \ \tilde{\tau} = 0.
\end{align}
\end{subequations}
This concludes that $\dot{V}(x)$ is always negative and therefore the \ac{ds} converges to the curve $\gamma(\tau)$ on the local domain $\mathcal{D}$.}

\subsubsection{\added{Cyclic motions}}
\added{For cyclic motions, as described in Sec.~\ref{sssec: cyclic_motions}, the \ac{ds} should converge to a closed curve, i.e. $\gamma(0) = \gamma(1)$. This simplifies the statements on convergence to the curve on the domain $\mathcal{D}$. In~\eqref{eq:proof_int_parts}, only Case 1 needs to be considered from the analysis as $\gamma'(\tau)$ and $\eta_x'(1)$ are strictly orthogonal on $\mathcal{D}$. For a \ac{ds} with a cyclic curve, the derivative of the Lyapunov function therefore becomes
\begin{equation}
\frac{dV(x)}{dt} = - 
k_{\text{NC}} d_M^2, \ \ \forall \ \tilde{\tau} \in [0, 1],
\end{equation}
which concludes converge to the cyclic curve $\gamma(\tau)$ on $\mathcal{D}$.}

\subsection{\added{Local convergence to the goal, $\gamma(1)$}} \label{ssec: part2_local_conv_goal}
\added{In addition to ensuring local convergence to the curve, we require the \ac{ds} to converge to the terminal point of the (non-cyclic) curve, namely the goal at $\gamma(1)$. To this end, within the local domain $\mathcal{D}$, we augment the Lyapunov candidate introduced in~\eqref{eq: lyapunov_conv_curve}, which guarantees local convergence to the curve, with the tangential potential
$\Psi_{\mathrm{TC}}(x)=\frac{1}{2}\left(1-\tilde{\tau}(x)\right)^2,$
which penalizes the remaining distance to the goal along the curve and has its minimum at $\tilde{\tau}(x)=1$. The resulting Lyapunov candidate is given by
\begin{equation}\label{eq:lyapunov_full}
\begin{aligned}
V(x)=\Psi_{\mathrm{NC}}(x)+\Psi_{\mathrm{TC}}(x).
\end{aligned}
\end{equation}
In the following, we show that this Lyapunov function establishes the local stability of $x^{*} = \gamma(1)$ within $\mathcal{D}$.}

\textbf{Positive Definiteness:}
By definition, the convergence term $\Psi_{\text{NC}}(x)$ is equal to zero on the curve, \ie $\Psi_{\text{NC}}(x) =0, \forall x \in \gamma$, and greater than zero when not on the curve, \ie $\Psi_{\text{NC}}(x) > 0, \forall x \notin\gamma$. The term $\Psi_{\text{TC}}(x)$ is nonnegative, with equality to zero iff $\tilde{\tau} = 1$. Thus, $V(x) \geq 0$, and $V(x) = 0$ iff $x = \gamma(1)$.

\textbf{Properness:} \added{We assume} that the sublevel sets of $V$ are compact. For non-compact manifolds like $SE(3)$, this property only holds when restricting the analysis to a bounded subset of the manifold $M$, which is a valid assumption for the workspace of a robotic manipulator.

\textbf{Dissipation:} We analyze the time derivative of $V(x)$ to establish convergence to the goal. Using $\dot{x}$ given by the DS in~\eqref{eq:ds} and the chain rule on the Riemannian manifold~\cite{bulloGeometricControlMechanical2005}:
\added{\begin{subequations} \label{eq: vdot_to_goal_start}
\begin{align}
    \dot{V} &= \dot{\Psi}_{\text{NC}} + \dot{\Psi}_{\text{TC}}\\[0.3em]
    &= -
\left\langle
\eta_x'(0),
\dot{x}
\right\rangle  + \langle  \nabla \Psi_{\text{TC}}, \dot{x} \rangle, \notag\\
&= -
\left\langle
\eta_x'(0),
\dot{x}
\right\rangle  + \langle  \nabla \Psi_{\text{TC}}, -k_{\text{NC}} \nabla \Psi_{\text{NC}} + k_{\text{TC}} \PT{\proj(x)}{x}(\gamma') \rangle, \notag\\
        &= \underbrace{- \left\langle
\eta_x'(0),
\dot{x}
\right\rangle}_{\text{Term A}} - k_{\text{NC}} \underbrace{\langle \nabla \Psi_{\text{TC}} , \nabla \Psi_{\text{NC}} \rangle}_{\text{Term B}} \notag\\
    &\qquad + k_{\text{TC}} \underbrace{\langle \nabla \Psi_{\text{TC}}, \PT{\proj(x)}{x}(\gamma')  \rangle}_{\text{Term C}}.
\end{align}
\end{subequations}}

\added{Leveraging the results from Sec.~\ref{ssec: part1_local_to_curve}, Term A is already established in~\eqref{eq: dvdt_curve_attract_conclude}.
To analyze term B and C, we first focus on convergence when $\tilde{\tau} = (0, 1]$, and continue with the more challenging case when $\tilde{\tau} = 0$.}

\added{\textbf{When} $\bm{\tilde{\tau} = (0, 1]}$, let us analyze the tangential gradient $\nabla \Psi_{\text{TC}} = -(1-\tilde{\tau}(x)) \nabla \tilde{\tau}$.
$\tilde{\tau}$ is positively correlated to the curve derivative $\gamma'$. 
We can therefore identify $\nabla \tilde{\tau}$ as $\frac{1}{\beta(x)} \PT{\proj(x)}{x}(\gamma')$ where $\beta(x)>0$ accounts for the metric distortion in the tubular neighborhood and its positivity ensures that motion along the propagated curve tangent corresponds to increasing phase, $\beta(x)>0$~\cite{carmoRiemannianGeometry1992, leeIntroductionRiemannianManifolds2018}.
The tangential gradient becomes
\begin{equation} \label{eq: psi_TC_with_beta}
    \nabla \Psi_{\text{TC}} = -\frac{1-\tilde{\tau}}{\beta(x)} \PT{\proj(x)}{x}(\gamma'), \ \ \forall \tilde{\tau} \in (0, 1).
\end{equation} 
Inserting Term A and~\eqref{eq: psi_TC_with_beta} into~\eqref{eq: vdot_to_goal_start} gives}
\begin{align} \label{eq:dissiation_vdot}
    \added{\small{\dot{V}}} 
        &\added{\small{= -k_{\text{NC}} d_{M}^2(\gamma, x) - k_{\text{NC}} \underbrace{\frac{-(1-\tilde{\tau})}{\beta(x)} \langle \PT{\proj(x)}{x}(\gamma'), \nabla \Psi_{\text{NC}} \rangle}_{\text{Term B}}}} \notag\\
    &\quad
        \added{\small{+ k_{\text{TC}} \underbrace{\frac{-(1-\tilde{\tau})}{\beta(x)} \langle \PT{\proj(x)}{x}(\gamma'), \PT{\proj(x)}{x}(\gamma') \rangle}_{\text{Term C}},  \ \ \forall \tilde{\tau} \in (0, 1).}}
\end{align}
By the First Variation of Arc Length~\cite{leeIntroductionRiemannianManifolds2018}, the geodesic radius vector, and thus $\nabla \Psi_{\text{NC}}$, is strictly orthogonal to the parallel transported tangent vector $\PT{\proj(x)}{x}(\gamma')$, for $x$ with $\tilde{\tau} \in (0, 1)$ on $\mathcal{D}$.
\added{Consequently, Term B, vanishes for $\tilde{\tau} \in (0, 1)$.}
For points $x$ projecting towards the end of the curve, $\tilde{\tau} = 1$, the factor in~\eqref{eq:twist_scaling_to_zero} sets $\gamma'(1)$ to zero, such that terms B and C vanish.
\added{Hence, for $\tilde{\tau} \in (0, 1]$ and $x \neq x^{*}$, \eqref{eq:dissiation_vdot}~simplifies to:
\begin{equation}\label{eq:v_dot_simplified}
    \dot{V}(x) = - k_{\text{NC}} d_M^2 - \frac{k_{\text{TC}}}{\beta(x)} (1 - \tilde{\tau}(x)) C < 0. 
\end{equation}
Since $\beta(x) > 0$ inside the tubular neighborhood, and $C := \langle \PT{\proj(x)}{x}(\gamma'), \PT{\proj(x)}{x}(\gamma') \rangle > 0$ for $\tilde{\tau} < 1$ and $C = 0$ for $\tilde{\tau} = 1$, both terms in~\eqref{eq:v_dot_simplified} are non-positive and vanish simultaneously only at the target state $x^{\ast} = \gamma(1)$.}

\added{\textbf{When $\bm{\tilde{\tau}} \bm{=} \bm{0}$}, the closest point projection lies at the beginning of the curve. 
As reasoned in Case 3 in App.~\ref{ssec: part1_local_to_curve}, when inside the local neighborhood where $\tilde{\tau} = 0$, the projected phase does not change, therefore $\nabla \tilde{\tau} = 0$. Inside the local neighborhood where $\tilde{\tau} = 0$, it therefore holds that $\nabla \Psi_{\text{TC}}=0$, and Term B and C vanish.
Inside this set, the derivative of the Lyapunov function is therefore decreasing, 
\begin{equation}
\dot{V} \leq - k_{\text{NC}} d_M^2 < 0, \ \text{if} \ \tilde{\tau} = 0.
\end{equation}
It remains to show that trajectories cannot remain in the local neighborhood where $\tilde{\tau} = 0$ indefinitely. 
Let $\mathcal{B} = \{x\in \mathcal{D} \ | \ \pi(x) = \gamma(0)\}$ denote the set of states whose closest-point projection coincides with the initial point of the curve. On the boundary of this region, points project orthogonally onto the curve. On the boundary, we can therefore use the same derivative of the Lyapunov candidate as in~\eqref{eq:v_dot_simplified}.
For points on $\mathcal{B}$ that lie on the interior of the set $\mathcal{B}$, the converging and propagating vectors in the \ac{ds} are not orthogonal, and the inner product is positive,
$\langle -\nabla \Psi_{\text{NC}}, \PT{\proj(x)}{x}(\gamma') \rangle > 0$.
The tangential component therefore possesses a positive component in the forward direction of the curve. 
Consequently, $\mathcal{B}$ is not positively invariant: trajectories cannot remain confined to $\mathcal{B}$, but are driven towards the interior region where $\tilde{\tau} \in (0, 1)$. Hence, there exists a finite time $t_{\mathcal{B}}>0$ such that
\begin{equation}
    \pi(x(t)) = \gamma(\tilde{\tau}(t)), \ 0 < \tilde{\tau}(t) < 1, \  \forall t > t_{\mathcal{B}}.
\end{equation}
Once the trajectory enters this region, the Lyapunov analysis of \eqref{eq:v_dot_simplified} applies. 
In conclusion, $\dot{V}(x)$ in~\eqref{eq:lyapunov_full} becomes}
{\small
\begin{subequations}
\begin{align}
\added{\dot{V}(x)}
&\added{= - k_{\text{NC}} d_{M}^2(\gamma, x)
- \frac{k_{\text{TC}}}{\beta(x)} (1 - \tilde{\tau}) C \leq 0,}
&& \added{\forall \tilde{\tau} \in (0, 1]}
\label{eq:lyapunov_full_a}\\
\added{\dot{V}(x)}
&\added{\leq - k_{\text{NC}} d_M^2 < 0,}
&& \added{\text{if } \tilde{\tau} = 0,}
\end{align}
\end{subequations}
}%
\added{with strict inequality of~\eqref{eq:lyapunov_full_a} when $x \neq x^{*}$. Therefore, all trajectories that start in $\mathcal{D}$ and stay in $\mathcal{D}$, converge to the equilibrium point $x^{*} = \gamma(1)$. }

\subsection{\added{Analyzing convergence in the presence of $\mathcal{C}_{\mathrm{sing}}$}} \label{ssec: part3_convergence_local_max}
\added{In App.~\ref{ssec: part1_local_to_curve} and~\ref{ssec: part2_local_conv_goal}, the assumption is made that the \ac{ds} operates on the domain $\mathcal{D}$, which excludes the singular set $\mathcal{C}_{\mathrm{sing}}$. 
As in App.~\ref{ssec: part1_local_to_curve} and~\ref{ssec: part2_local_conv_goal}, trajectories are assumed to not start in $\mathcal{C}_{\mathrm{sing}}$, as this set is of measure zero, and no trajectories convergence to $\mathcal{C}_{\mathrm{sing}}$, since $\mathcal{C}_{\mathrm{sing}}$ contains unstable equilibria of the distance potential. 
However, as observed in Fig.~\ref{fig:lasa_ridges}, trajectories may \textit{pass through} $\mathcal{C}_{\mathrm{sing}}$.

For points in $\mathcal{C}_{\mathrm{sing}}$, two different cases can be established, denoted by
$\mathcal{C}_{\mathrm{sing}}
=
\left\{
\mathcal{C}_{\mathrm{sing},1},
\mathcal{C}_{\mathrm{sing},2}
\right\}$:}
\begin{itemize}
    \item \added{The set $\mathcal{C}_{\text{sing,1}}$ contains points $x$ that have a minimum distance to multiple (distinct) points on the curve.}
    \item\added{The set $\mathcal{C}_{\text{sing,2}}$ contains points $x$ that have multiple geodesics to the same point on the curve.}
\end{itemize}

\added{For $\mathcal{C}_{\text{sing,2}}$, computing the set of geodesics that all are valid shortest paths to the curve, as well as knowing when a trajectory passes through $\mathcal{C}_{\text{sing,2}}$, can be challenging on Riemannian manifolds and Lie groups. Therefore, we assume that on $\mathcal{C}_{\text{sing,2}}$, the selected geodesic defines a valid shortest path and velocity vector that ensures trajectories passing through $\mathcal{C}_{\text{sing,2}}$ enter $\mathcal{D}$. Practically, in Sec.~\ref{sec:experimental_results}, convergence was achieved across all experiments.
The remaining analysis therefore focuses on crossings of $\mathcal{C}_{\text{sing,1}}$, where multiple closest-point projections may require explicit treatment.
}

\subsubsection{Passing through $\mathcal{C}_{\text{sing,1}}$}
\added{Given \(x\in\mathcal{C}_{\mathrm{sing,1}}\), let \(\mathcal{P}_{\tau}(x)\) denote the non-empty compact set of phase values corresponding to
these closest points on the curve:
\begin{equation}
\mathcal{P}_{\tau}(x)
=
\operatorname*{arg\,min}_{\tau\in[0,1]}
d_{M}(x,\gamma(\tau)).
\end{equation}
Consequently, a maximal phase value exists,
\begin{equation} \label{eq:high_phase_passing}
    \tau^{\ast}
    =
    \max_{\tau\in\mathcal{P}_{\tau}(x)} \tau .
\end{equation}
We resolve the ambiguity of the projection map by selecting the closest point with the largest phase value, thereby choosing the projection furthest along the reference curve.}

\added{
The vector field generated by $f(x)$ is piecewise smooth, with possible discontinuities on $\mathcal{C}_{\mathrm{sing,1}}$~\cite{bernardo2008piecewise}. 
We therefore analyze the energy function separately on the regular regions and establish a non-increase condition when a trajectory crosses the singular ridge~\cite{bernardo2008piecewise, branicky1998multiple}.
On $\mathcal{D}$, $V(x)$ is continuously differentiable and strictly decreasing, as in App.~\ref{ssec: part2_local_conv_goal}. It therefore remains to show that the energy does not increase when a trajectory crosses $\mathcal{C}_{\mathrm{sing,1}}$. Assume that each trajectory intersects $\mathcal{C}_{\mathrm{sing,1}}$ only finitely many times. In addition, we know that the curve is contained in the regular domain.}

\added{Consider a crossing from a regular region $\mathcal D_1$ through $\mathcal{C}_{\mathrm{sing,1}}$ into a regular region $\mathcal D_2$, where $\mathcal{D}_1\ \cap \ \mathcal{C}_{\mathrm{sing,1}}=\emptyset$ and $\mathcal{D}_2 \cap \ \mathcal{C}_{\mathrm{sing,1}}=\emptyset$.
The transition from $\mathcal{C}_{\mathrm{sing,1}}$ to $\mathcal D_2$ is $C^0$-continuous under~\eqref{eq:high_phase_passing}, and hence the decreasing energy $V(x)$ of App.~\ref{ssec: part2_local_conv_goal} remains valid on $\mathcal{C}_{\mathrm{sing,1}} \cup \mathcal{D}_{2}$. The potentially discontinuous transition is therefore the one from $\mathcal D_1$ to $\mathcal{C}_{\mathrm{sing,1}}$. It remains to show that the energy does not increase at this transition.
The maximal-phase selection favors the projection furthest along the reference curve and therefore ensures}
\added{\begin{equation} \label{eq: tau_singular_forward_pass}
\tilde{\tau}(x_{\mathcal{C}_{\text{sing, 1}}}) \geq \tilde{\tau}(x_{\mathcal{D}_1}).
\end{equation}

Consider separately the normal and tangential terms of $V(x)$. For $\Psi_{\mathrm{NC}}(x)$, the distance to the curve is $C^0$-continuous when entering $\mathcal{C}_{\text{sing,1}}$ from $\mathcal{D}_1$. Moreover, along the crossing, it holds that $\langle -\nabla \Psi_{\text{NC}}, \PT{\proj(x)}{x}(\gamma') \rangle \geq 0$. Consequently, $\Psi_{\mathrm{NC}}$ does not increase at the crossing, yielding $\Psi_{\mathrm{NC}}(x_{\mathcal{C}_{\text{sing,1}}}) \leq \Psi_{\mathrm{NC}}(x_{\mathcal{D}_1})$. 
For $\Psi_{\mathrm{TC}}(x)=\frac{1}{2}\left(1-\tilde{\tau}(x)\right)^2$, given the non-decreasing projected phase over the crossing in~\eqref{eq: tau_singular_forward_pass}, it follows that 
$\Psi_{\mathrm{TC}}(x_{\mathcal{C}_{\text{sing,1}}}) - \Psi_{\mathrm{TC}}(x_{{\mathcal{D}_1}}) \leq 0$.
Combining the two components gives, $V(x_{\mathcal{C}_\mathrm{sing,1}}) - V(x_{\mathcal{D}_1}) \leq 0$, thus $V$ is non-increasing at every crossing of $\mathcal{C}_\mathrm{sing,1}$.}

\added{Since $V$ strictly decreases during the continuous evolution within each regular region and does not increase at the finitely many crossings, $V$ is non-increasing along the entire trajectory. After the final crossing, the trajectory remains in a regular region and the convergence result of App.~\ref{ssec: part2_local_conv_goal} applies. Therefore, the trajectory converges to the goal $x^{*} = \gamma(1)$.}

\added{Note the similarity of our analysis with the hybrid feedback formulation in~\cite{mayhew2013synergistic}, where the active potential is selected from a finite family according to their energy values and switching logic, to avoid undesired critical points. Similarly, our \ac{ds} selects the minimum energy function by selecting $\tilde{\tau}$. However, whereas the family of potential functions in~\cite{mayhew2013synergistic} is finite, the corresponding family $\mathcal{P}_{\tau}$ in our formulation may be infinite.}